\documentclass[11pt]{article}

\usepackage[utf8]{inputenc}
\usepackage[T1]{fontenc}
\usepackage[a4paper,margin=1in]{geometry}
\usepackage{amsmath,amssymb,amsfonts,bm}
\usepackage{graphicx}
\usepackage{booktabs}
\usepackage{array}
\usepackage{caption}
\usepackage{subcaption}
\usepackage[protrusion=false,expansion=false]{microtype}
\usepackage[round,authoryear]{natbib}
\usepackage[hidelinks]{hyperref}
\usepackage{setspace}
\usepackage{float}
\usepackage{authblk}
\usepackage{hyperref}

\graphicspath{{figures/}}

\newcommand{\Lset}{\mathcal{L}}
\newcommand{\Uset}{\mathcal{U}}
\newcommand{\Xset}{\mathcal{X}}
\newcommand{\Sset}{\mathcal{S}}
\newcommand{\Hk}{\mathcal{H}_k}
\newcommand{\pl}{P_{S}}
\newcommand{\pu}{P_{U}}

\newcommand{\MMD}{\mathrm{MMD}}
\newcommand{\KL}{D_{\mathrm{KL}}}
\newcommand{\Wass}{W_{1}}
\newcommand{\neff}{n_{\mathrm{eff}}}
\newcommand{\Real}{\mathbb{R}}
\newcommand{\E}{\mathbb{E}}
\newcommand{\indic}[1]{\mathbb{1}\{#1\}}

\title{\bf Semi-Supervised Learning under Spatially Biased Sampling}

\author[1]{Bright Wiredu Nuakoh}
\author[2,3,4]{Francky Fouedjio}
\author[3]{Stephen Bradshaw}
\author[2,5]{Yaw Kwaafo Awuah-Mensah}
\author[3]{Wei Hong Tan}
\author[2,3]{Emet Arya}
\author[2,5]{Ebenezer Afrifa-Yamoah}

\affil[1]{African Institute for Mathematical Sciences (AIMS), 17 KN 16 Ave, Kigali, Rwanda}
\affil[2]{Mathematical Application \& Data Analytics Group, School of Science, Edith Cowan University, 270 Joondalup Drive, Joondalup, WA 6027, Australia}
\affil[3]{Data \& Analytics, Rio Tinto, 152-158 St Georges Terrace, Perth, WA 6000, Australia}
\affil[4]{Kaplan Business School Pty Ltd, Perth Campus, 1325 Hay St, West Perth, WA, 6005, Australia}
\affil[5]{Centre for Marine Ecosystems Research, School of Science, Edith Cowan University, 270 Joondalup Drive, Joondalup, WA 6027, Australia}

\affil[ ]{\textbf{Corresponding author:} E Afrifa-Yamoah (\href{mailto:e.afrifayamoah@ecu.edu.au}{e.afrifayamoah@ecu.edu.au})}
\date{\today}

\begin{document}
\maketitle

\begin{abstract}
\noindent
Standard semi-supervised learning (SSL) typically relies on labelled and unlabelled data sharing a common marginal distribution. This assumption is often violated by biased spatial  sampling mechanism, when labels are collected under spatially biased or preferential site selection. We treat this marginal mismatch, spatial autocorrelation, and spatial non-stationarity as three distinct mechanisms, varied independently via a labelled-sampling concentration parameter, a spatial length scale, and a non-stationarity strength parameter, and ask how mismatch degrades SSL, whether the cluster and manifold assumptions survive it, and how the resulting failure can be diagnosed. Using a controlled synthetic framework alongside PovertyMap-WILDS, California housing, socio-economic and US air quality monitoring datasets, we systematically vary the degree of mismatch while accounting for spatial autocorrelation and non-stationarity. Through a series of analyses including a segmented-regression changepoint, we show that in the synthetic generator, SSL performance does not degrade gradually but instead exhibits a threshold-like breakdown between approximately 0.71 and 0.77 once distribution mismatch becomes sufficiently severe. We further demonstrate that spatial non-stationarity contributes to performance loss independently of marginal mismatch and that models become increasingly overconfident outside the regions where labels are available. To support practical deployment, we evaluate several distribution-divergence measures as indicators of reliability and introduce a kernel-weighted local divergence metric that provides a more stable estimate of spatial mismatch than a naïve localised approach. These findings provide empirical evidence and diagnostic tools for better documenting the risk of incorporating unlabelled spatial data into semi-supervised learning workflows.\\

Keywords: marginal distributional mismatch, preferential sampling, spatial sampling, machine learning, AI

\end{abstract}

\section{Introduction}
 
Semi-supervised learning (SSL) utilises abundant unlabelled data to compensate for scarce, costly labels \citep{chapelle2006ssl, vanengelen2020survey}. Its guarantees rest on a distributional-consistency assumption: that labelled and unlabelled samples are drawn from the same distribution. Spatial data can seem to inherently violate this assumption, but that
framing is imprecise. Labelled and unlabelled locations are ordinarily realisations of the
same spatial stochastic process, and the defining features of such a process -- spatial autocorrelation, whereby nearby locations are more similar than distant ones
\citep{tobler1970computer}, and spatial non-stationarity, whereby the covariate -- outcome relationship varies across geography \citep{fotheringham2002gwr} -- do not by themselves make the labelled and unlabelled marginals differ. A second-order non-stationary Gaussian random
field, for instance, exhibits both properties, yet observations at labelled and unlabelled locations remain realisations of one process and can be modelled as such
\citep{banerjee2014hierarchical, paciorek2006spatial}. What violates the shared-marginal assumption is the sampling mechanism: labelled spatial data are collected under practical constraints where survey budgets, monitoring-station siting, and historical settlement patterns concentrate labels in accessible or well-resourced areas, so that the labelled and unlabelled covariate marginals diverge through spatially biased coverage, preferential sampling in which site selection depends on the process being measured
\citep{diggle2010preferential}, or domain shift between collection regimes. Autocorrelation and non-stationarity are nonetheless consequential for SSL, but through different channels: autocorrelation stresses the independence assumptions behind SSL's learning-theoretic guarantees \citep{roberts2017crossval}, while non-stationarity is a form of concept shift in $p(y\mid x)$ \citep{morenotorres2012unifying}. Whether these three mechanisms degrade SSL in the same way, to the same degree, and through the same failure modes has not been established, even as the wider literature documents how poorly spatial models generalise beyond their training footprint \citep{meyer2021aoa, ploton2020spatial}.

The paper makes three primary contributions\textsuperscript{\textdagger}. First, it frames marginal mismatch and spatial
non-stationarity as distinct forms of dataset shift within the taxonomy of \citet{morenotorres2012unifying}, and provides empirical evidence that they affect SSL performance through different mechanisms in the settings studied. Second, it evaluates global distribution-divergence metrics as practical indicators of SSL reliability and introduces a kernel-weighted local divergence estimator that provides a more robust measure of spatial mismatch than a naïve grid-based approach. Third, it shows that SSL performance exhibits a threshold-like breakdown under increasing spatial mismatch and estimates this transition using segmented regression with bootstrap confidence intervals \citep{muggeo2003segmented}. Under the experimental settings considered here, the estimated breakpoint lies at approximately $\alpha \approx 0.71$--$0.77$ for five of the six method/metric combinations tested (the sixth, label propagation's accuracy breakpoint, is far less precisely located), indicating that degradation is abrupt instead of gradual. Beyond these core contributions, we also investigate two related questions: whether the observed failure can be mitigated through distribution-aware training, and whether the same behaviour is consistent across a broader range of SSL methods. These analyses include a distribution-aware framework, a domain-adversarial variant, and a comparison of ten methods spanning four SSL families, including the geostatistical approach of \citet{fouedjio2022geostatistical}. Unless otherwise stated, all results presented in this paper are based on twenty random seeds. As these investigations extend the main narrative of the paper, they are summarised briefly in the main text and presented in full in the Supplementary Material.
{\let\thefootnote\relax\footnotetext{$^\dagger$Code available at \url{https://github.com/Wiredu2020/Distributional-Mismatch-in-Spatial-SSL-Application}}}

The study also tests the validity of the following hypotheses, which are derived from the research questions we seek to address. The mismatch-degradation hypothesis (H1) states that increasing marginal distribution mismatch between labelled and unlabelled spatial data degrades SSL model performance. The method-sensitivity hypothesis (H2) states that methods relying strongly on cluster or manifold assumptions are more sensitive to spatial heterogeneity than methods that do not. The spatial-awareness hypothesis (H3) states that spatially explicit SSL models outperform standard, non-spatial SSL approaches under distribution mismatch. The alignment-benefit hypothesis (H4) states that incorporating distribution-alignment or spatial-weighting mechanisms directly into the training procedure mitigates the performance loss predicted by the mismatch-degradation hypothesis.
\section{Related Work}
 
SSL families are organised by the structural assumption each operationalises
\citep{chapelle2006ssl, vanengelen2020survey}. The cluster assumption, that points in
the same high-density region share a label, underlies self-training and
pseudo-labelling \citep{lee2013pseudo}; the manifold assumption, that labels vary
smoothly along lower-dimensional manifolds, is realised by graph methods such as label
propagation \citep{zhu2002learning} and graph convolutional networks
\citep{kipf2017gcn}; and consistency regularisation requires stable predictions under
perturbation, as in Mean Teacher \citep{tarvainen2017mean}, MixMatch
\citep{berthelot2019mixmatch}, and FixMatch \citep{sohn2020fixmatch}. Each mechanism
depends on the unlabelled data resembling the evaluation distribution.
\citet{oliver2018realistic} showed this is not a technicality: performance falls once
the unlabelled pool contains out-of-distribution observations, so matched-pool
benchmarks overstate reliability. That study, however, constructed shift by withholding
classes instead of varying where in space data were collected, leaving open whether
the same degradation arises when mismatch is geographic. This concern has also motivated
safe semi-supervised learning, which asks whether using unlabelled data can be prevented
from performing worse than the corresponding supervised learner \citep{lizhou2015safe}.

Two properties of spatial data are central here but rarely treated in SSL.
Autocorrelation, Tobler's first law \citep{tobler1970computer, legendre1993spatial},
violates the independence assumed by learning-theoretic guarantees;
\citet{roberts2017crossval} show that ignoring it, for instance through random rather
than spatially blocked cross-validation, inflates apparent performance, though the
appropriate design remains debated \citep{wadoux2021spatial}. Non-stationarity, the
variation of the covariate--outcome relationship across space, motivated geographically
weighted regression \citep{fotheringham2002gwr} and, more formally, spatially varying
coefficient processes \citep{gelfand2003spatially}. These are distinct from one another and
from marginal mismatch: autocorrelation concerns how covariates co-vary in space,
non-stationarity how the covariate--label map varies in space, and mismatch where the
samples were drawn from. How they interact for SSL, and in particular whether
non-stationarity degrades performance independently of mismatch, has not to our
knowledge been established empirically.
 
Marginal mismatch is an instance of covariate shift within the dataset-shift taxonomy
of \citet{morenotorres2012unifying}, which separates it from prior-probability and
concept shift \citep[]{quinonero2009dataset}. \citet{shimodaira2000improving}
showed that estimation is biased under covariate shift and proposed importance-ratio
re-weighting, the basis for much of domain adaptation \citep{pan2010survey,
sugiyama2008direct, sugiyama2012machine, ganin2016domain}. Domain-adaptation theory shows
that target error depends on source error, domain discrepancy, and the existence of a
predictor that performs well in both domains, so a small discrepancy alone is not a
sufficient safety guarantee \citep{bendavid2010theory}; relatedly, domain-invariant
representations can still fail when the source and target conditional or label
distributions differ, so marginal alignment alone is not a complete solution
\citep{zhao2019invariant}. The degree of mismatch
is quantified by kernel two-sample tests such as maximum mean discrepancy
\citep{gretton2012kernel} and optimal-transport distances such as the Wasserstein metric
\citep{courty2016optimal}. What this literature leaves open is the interaction with
spatial structure: a globally computed divergence can obscure localised mismatch, while a
naïvely localised one becomes unreliable where local samples are too small, a
bias--variance tension spatial extensions have only begun to confront. Domain adaptation
has a long history in remote sensing specifically, where classifiers trained in one
region, sensor, or time period may not transfer reliably to another \citep{tuia2016domain}. The WILDS suite provides real-world geographic shift tasks such as satellite poverty mapping \citep{koh2021wilds}, but treats shift as a fixed in-distribution/out-of-distribution split rather than a quantity to be varied, and does not compare SSL families. This paper uses PovertyMap-WILDS as a real-data counterpart to the synthetic mismatch sweep, restricting the labelled countries to induce increasing severity. A separate strand builds spatial structure into learning: geographically neural network weighted regression learns geographic weights within a locally weighted regression model \citep{du2020gnnwr}, an
approach later applied to house-price valuation \citep{wang2022gnnwr}, and, closest to the present
work, \citet{fouedjio2022geostatistical} generate pseudo-labels at unlabelled sites by
geostatistical conditional simulation, building on model-based geostatistics for
non-Gaussian outcomes \citep{diggle1998modelbased} and, for categorical outcomes,
indicator geostatistics \citep{journel1983nonparametric,chiles2012geostatistics,
goovaerts1997geostatistics}; they outperform self-training on continuous geochemical targets, though without a controlled mismatch sweep or a comparison against graph- or consistency-based SSL.
 
The stakes are highest where SSL is most attractive because labels are costly. Housing
valuation labels track observed transactions clustered in active markets, motivating
geographically weighted and neural hedonic models \citep{wang2022gnnwr}; the California
Housing data used here originate from this literature \citep{pace1997sparse}.
Socio-economic indicators such as poverty are sparsely surveyed, motivating the
combination of limited survey labels with dense unlabelled covariates such as satellite
imagery \citep{jean2016combining, rolf2021generalizable, usda2024county}. Air-quality
monitors are non-randomly sited, forcing land-use regression to extrapolate from a
spatially biased sample \citep{hoek2008lur, epa2024aqs}.
 
\section{Problem Formulation and Methods}
 
\subsection{Notation and probabilistic formulation}
\label{sec:setup}

Let $\{Z(s) = (X(s), Y(s)) : s \in \Sset\}$, $\Sset \subset \Real^{2}$, be a spatial
stochastic process, with covariate field $X(s) \in \Xset \subseteq \Real^{d}$ and binary
response $Y(s) \in \{0,1\}$. The process is characterised by a covariance structure
$C(s,s') = \operatorname{Cov}\!\big(X(s), X(s')\big)$, which encodes spatial
autocorrelation and is second-order stationary when $C$ depends only on the lag $s-s'$, and
by a conditional response law $\eta_s(x) = \Pr\!\big(Y(s)=1 \mid X(s)=x\big)$, whose variation
with $s$ constitutes spatial non-stationarity. Autocorrelation and
non-stationarity are properties of this single process: observations at any set of locations
are realisations of the same $Z(\cdot)$, and can be represented within standard geostatistical
and hierarchical spatial frameworks \citep{cressie1993statistics, banerjee2014hierarchical,
paciorek2006spatial}.

Learning proceeds from a labelled set $\Lset = \{(s_i, x_i, y_i)\}_{i=1}^{n_\ell}$, drawn from
a source distribution $P_S$, and an unlabelled set $\Uset = \{(s_j, x_j)\}_{j=1}^{n_u}$,
$n_u \gg n_\ell$, drawn from $P_U$; a third, generally unobserved distribution $P_T(x,y)$
governs deployment -- the region a fitted model is actually evaluated against, realised here
by the out-of-region test set. Their locations are drawn under sampling designs with spatial
densities $\pi_\ell(s)$, $\pi_u(s)$, and $\pi_t(s)$ respectively. These induce covariate
marginals over the sampled locations,
\begin{equation}
\pl(x) \;\propto\; \int_{\Sset} f_{X(s)}(x)\, \pi_\ell(s)\, ds, \qquad
\pu(x) \;\propto\; \int_{\Sset} f_{X(s)}(x)\, \pi_u(s)\, ds,
\label{eq:induced-marginals}
\end{equation}
where $f_{X(s)}$ is the marginal density of the field at $s$; $P_T$ is defined analogously
from $\pi_t(s)$, over the same conditional law $\eta_s$, so $P_T(x,y) = \int_{\Sset}
f_{X(s)}(x)\,\eta_s(x)^{y}(1-\eta_s(x))^{1-y}\,\pi_t(s)\,ds$. Semi-supervised learning rests
on two assumptions. The first is covariate consistency: the labelled and unlabelled
inputs share a marginal,
\begin{equation}
\text{(A1)}\qquad \pl(x) = \pu(x). \label{eq:covshift}
\end{equation}
The second is a structural assumption on the conditional -- the cluster, manifold, or
low-density separation conditions -- under which $p(y\mid x)$ is smooth enough that unlabelled
covariates are informative about the decision boundary. Note that $\pi_u$ and $\pi_t$ are set independently throughout (uniform over the
domain, and $P_U = P_T$ by construction, but varied against each other), thus, the unlabelled pool is drawn from the training regions while the out-of-region test
set is held out from different regions entirely, resulting in $P_U \neq P_T$ in general. The goal is a
predictor $f:\Xset\to[0,1]$ with small target risk
$R_{T}(f) = \E_{(x,y)\sim P_T}\big[\ell(f(x),y)\big]$, realised in-sample by the out-of-region
test set; we track the generalisation gap $\Delta = \mathrm{Acc}_{\mathrm{in}} -
\mathrm{Acc}_{\mathrm{out}}$ as a finite-sample proxy for $R_{\mathrm{in}}(f) - R_T(f)$. 

\begin{figure}[H]
  \centering
  \includegraphics[width=\linewidth]{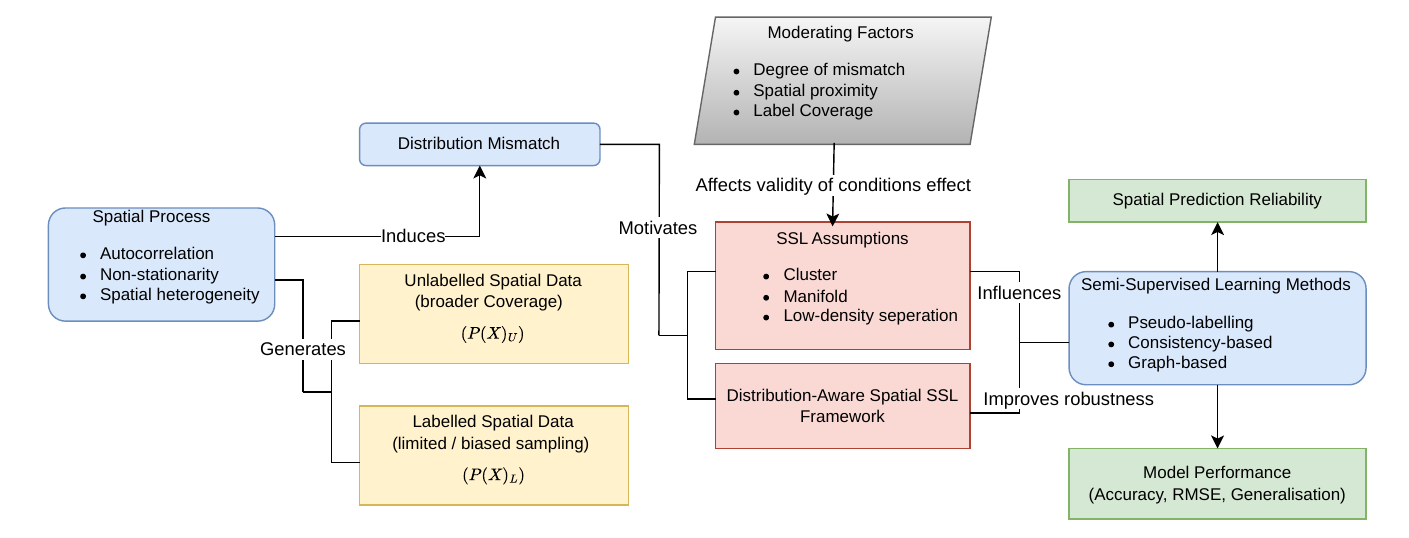}
  \caption{Conceptual framework linking marginal mismatch between labelled and
  unlabelled spatial data to the validity of semi-supervised learning assumptions and
  to model performance. Marginal mismatch arises from the sampling mechanism, while distribution-aware approaches
  aim to mitigate its effect.}
  \label{fig:framework}
\end{figure}

By \eqref{eq:induced-marginals}, (A1) is a statement about the sampling design: it holds
whenever $\pi_\ell$ and $\pi_u$ give matched coverage of the field, and it does so even
when the field is strongly autocorrelated and non-stationary. Assumption (A1) is violated not
by spatial dependence but by the data-generating and sampling mechanism, in one of three ways:
(i) spatially biased coverage, $\pi_\ell \neq \pi_u$ over a heterogeneous field, which shifts
$p(x)$ while leaving $p(y\mid x)$ fixed -- covariate shift in the sense of
\citet{morenotorres2012unifying} and \citet{shimodaira2000improving}; (ii) preferential
sampling, in which $\pi_\ell$ depends on the latent process $Z$ itself \citep{diggle2010preferential};
or (iii) domain shift, in which $\Lset$ and $\Uset$ arise under different process laws.
Spatial non-stationarity is a separate departure --- a variation of $\eta_s = p(y\mid x, s)$
with location, i.e. concept shift ---
\begin{equation}
\text{(non-stationarity, concept shift):}\qquad p(y\mid x, s)\ \text{varies with}\ s,
\label{eq:concshift}
\end{equation}
which is orthogonal to (A1) and can hold or fail independently of it. This paper studies a
controlled departure from (A1) driven by the sampling design, holding the field fixed,
and treats autocorrelation and non-stationarity as distinct mechanisms, not as causes of
marginal mismatch. Figure~\ref{fig:framework} summarises how the
three spatial mechanisms feed distributional divergence, the SSL assumptions, and
downstream prediction reliability.
 
\subsection{Synthetic data-generating process}
\label{sec:synthetic}
 
The generator exposes three independent axes, each expressed as a standard construct in
spatial statistics, so that the mechanisms of Section~\ref{sec:setup} can be varied in
isolation. The covariate field $X(s)$ is a Gaussian random field with a stationary covariance
kernel whose range parameter $\ell$ sets the strength of spatial autocorrelation
\citep{cressie1993statistics, chiles2012geostatistics}. Spatial non-stationarity is
introduced through spatially varying regression coefficients in the response law, in the manner
of geographically weighted models \citep{fotheringham2002gwr} and non-stationary covariance
constructions \citep{paciorek2006spatial, sampson1992nonparametric},
\begin{equation}
Y(s) = \indic{\beta(s)^{\top} X(s) + \varepsilon(s) > 0}, \qquad
\beta(s) = \big(\beta_0 + \tau\, g(s),\, \beta_1,\, \ldots,\, \beta_p\big)^{\top},
\end{equation}
with $X(s) \in \Real^{p}$ , $g(s) = \sin(2\pi s_1)\cos(2\pi s_2)$ a deterministic
field with zero mean over $\Sset$ and independent of $X(s)$, $\tau \ge 0$ the non-stationarity
strength, and $\varepsilon(s) \sim \mathcal{N}(0,
\sigma_\varepsilon^2)$ i.i.d.\ noise. At $\tau=0$ the conditional
law $\eta_s$ is constant across space. The field and its response
law are held fixed while the sampling design is varied: unlabelled and test locations are
drawn uniformly, whereas labelled locations follow a location density that concentrates toward
an anchor $s_0$ as $\alpha$ grows,
\begin{equation}
\pi_\ell^{(\alpha)}(s) \;\propto\; \exp\!\left(-\frac{\|s - s_0\|^{2}}{2\,h(\alpha)^{2}}\right),
\qquad h(\alpha) = (1-\alpha)\,h_{\max} + \alpha\,h_{\min},
\end{equation}
normalised to a proper probability mass function over the candidate locations. $\alpha = 0$ gives a
kernel wide enough to be near-uniform over $\Sset$ (assumption (A1) approximately satisfied)
and $\alpha = 1$ concentrates labels tightly around the anchor $s_0$. This design is deliberately non-preferential in
the sense of \citet{diggle2010preferential}: the location density $\pi_\ell^{(\alpha)}$ depends
only on geographic position, not on the latent response $Y(s)$, so the induced mismatch is
covariate shift arising from biased spatial coverage, not from stochastic dependence
between the sampling and the process. While $\ell$, $\tau$, and $\alpha$ enter as independent
parameters, the generator instantiates autocorrelation, concept shift, and marginal mismatch
separately, matching the conceptual separation of Section~\ref{sec:setup}.
 
\subsection{Measuring distribution mismatch}
\label{sec:divergence}
 
Three global divergences between $\pl(x)$ and $\pu(x)$ are used throughout. The
Kullback--Leibler divergence, with densities estimated by Gaussian kernel density estimation
using Scott's rule bandwidth \citep{silverman1986density} directly on the $p=2$ raw
covariates (no dimensionality reduction), is $\KL(\pl \,\|\, \pu) = \int
\pl(x)\,\log\{\pl(x)/\pu(x)\}\,dx$, approximated by a plug-in Monte Carlo average over the
labelled sample. This
one-directional estimate of $\KL(\pl\,\|\,\pu)$ can understate divergence when $\pu$ places
mass outside the support of $\pl$. The Wasserstein-1 distance between marginals is $\Wass(\pl,\pu) = \inf_{\gamma \in
\Pi(\pl,\pu)} \int \|x-x'\|\,d\gamma(x,x')$ \citep{courty2016optimal}, computed here as the
mean of per-feature one-dimensional Wasserstein distances on standardised covariates, following standard computational optimal-transport
reporting practice \citep{peyre2019computational}. The maximum mean
discrepancy \citep{gretton2012kernel}, with characteristic radial-basis-function kernel
$k$ and feature map $\phi$ satisfying $k(x,x')=\langle \phi(x),\phi(x')\rangle_{\Hk}$, is
\begin{equation}
\MMD^{2}(\pl,\pu) = \big\| \mu_{\pl} - \mu_{\pu} \big\|_{\Hk}^{2}
 = \E[k(x,x')] + \E[k(z,z')] - 2\,\E[k(x,z)],
\label{eq:mmd}
\end{equation}
with $x,x'\sim\pl$ and $z,z'\sim\pu$, estimated by the biased
plug-in ($V$-statistic) form, with RBF kernel $k(x,x') =
\exp(-\gamma\|x-x'\|^2)$ and bandwidth set by the median heuristic, $\gamma = 1/(2\tilde{d}^2)$, where $\tilde d$ is the median pairwise Euclidean distance in the pooled sample.
 
A spatially localised divergence should reveal where mismatch concentrates, not
only how much of it there is. The naïve construction partitions $\Sset$ into a grid
$\{C_g\}$, computes $\widehat{\MMD}^{2}_g$ within each cell retaining at least $m$
points from both samples, and averages over the qualifying set $\mathcal{G}_m$. This
construction fails under severe mismatch: as $\alpha \to 1$ local overlap collapses,
$|\mathcal{G}_m|$ shrinks, and the estimate becomes not merely less informative but dominated by sampling noise as demonstrated in (Section~\ref{sec:res-div}).

Hard cell membership was replaced with continuous spatial weighting. At each of
$T$ query locations $\{q_t\}$ spanning the domain, every
labelled and unlabelled point contributes a weight $w_i(q_t) = K_h(\|s_i - q_t\|)$ from a
Gaussian spatial kernel of fixed bandwidth $h$, instead of contributing only if it falls inside a hard cell. Using weighted, normalised kernel mean embeddings, the local squared
discrepancy at $q_t$ is
\begin{equation}
\widehat{\MMD}^{2}_h(q_t) =
  \left\| \frac{\sum_{i \in \Lset} w_i(q_t)\,\phi(x_i)}{\sum_{i \in \Lset} w_i(q_t)}
        - \frac{\sum_{j \in \Uset} w_j(q_t)\,\phi(x_j)}{\sum_{j \in \Uset} w_j(q_t)}
  \right\|_{\Hk}^{2},
\label{eq:localmmd}
\end{equation}
where the feature kernel $\phi$ uses the same characteristic RBF form and median-heuristic
bandwidth as the global MMD above. Each query location is summarised by two effective local sample sizes, one per
sample, the kernel-weighted analogue of a raw count obtained from the standard correction for
weighted samples \citep{kish1965survey},
\begin{equation}
\neff{}_{,\Lset}(q_t) = \frac{\big(\sum_{i \in \Lset} w_i(q_t)\big)^{2}}{\sum_{i \in \Lset} w_i(q_t)^{2}},
\qquad
\neff{}_{,\Uset}(q_t) = \frac{\big(\sum_{j \in \Uset} w_j(q_t)\big)^{2}}{\sum_{j \in \Uset} w_j(q_t)^{2}},
\label{eq:ess}
\end{equation}
and the domain aggregate weights each location by the more restrictive of the two,
\begin{equation}
D_{\mathrm{loc}} =
  \frac{\sum_{t=1}^{T} \min\!\big(\neff{}_{,\Lset}(q_t),\,\neff{}_{,\Uset}(q_t)\big)\, \widehat{\MMD}^{2}_h(q_t)}
       {\sum_{t=1}^{T} \min\!\big(\neff{}_{,\Lset}(q_t),\,\neff{}_{,\Uset}(q_t)\big) + \epsilon},
\label{eq:aggloc}
\end{equation}
Locations with little local support contribute correspondingly little to
\eqref{eq:aggloc}, instead of being included or excluded all at once. The construction
recasts spatial localisation as an estimation problem with an explicit bias--variance
trade-off instead of a binning exercise.
 
\subsection{Semi-supervised learning methods}
\label{sec:methods}
 
The main analysis contrasts three classical baselines that operationalise the cluster and
manifold assumptions directly. A further seven methods, spanning consistency-based,
graph-based, spatially explicit, and geostatistical families, together with the
distribution-aware framework introduced below, are evaluated under the identical protocol
and reported in full in the Supplementary Material while summarised in Sections~\ref{sec:res-framework}
and~\ref{sec:res-comparative}. A supervised-only control ignores $\Uset$ entirely.
Self-training \citep{lee2013pseudo} initialises $f^{(0)}$ on $\Lset$ using a
probability-calibrated support-vector base classifier and, at each round, assigns
pseudo-labels $\hat y_j = \arg\max_c f^{(t)}_c(x_j)$ to unlabelled points whose confidence
$\max_c f^{(t)}_c(x_j)$ exceeds a threshold $\tau$, augments the training set, and
refits.
Label propagation \citep{zhu2002learning} builds a RBF
affinity $W \in \Real^{N\times N}$, $W_{ij} = \exp(-\|x_i - x_j\|^{2}/2\sigma^{2})$, normalises it as $S = D^{-1/2}
W D^{-1/2} \in \Real^{N\times N}$, and diffuses labels through
$F^{(t+1)} = \zeta S F^{(t)} + (1-\zeta) Y$, which converges to the closed form
$F^{\star} = (1-\zeta)(I - \zeta S)^{-1} Y$.
 
\subsection{Distribution-aware framework}
\label{sec:framework}

To test H4 we layer a distribution-aware procedure on self-training. The first
component, density-ratio re-weighting, up-weights labelled points resembling the unlabelled
distribution $P_U$ -- the best available proxy for $P_T$, since the target itself is
unobserved at training time -- through an importance ratio $r(x)=\pu(x)/\pl(x)$ estimated
discriminatively,
in the spirit of the covariate-shift correction of \citet{shimodaira2000improving} and
\citet{sugiyama2008direct}; evaluated on its own as a fourth baseline, denoting re-weighted
self-training.

We initially added two further components, specified in full in Supplementary Material
Section~S2.2: a fixed-bandwidth geographic-proximity weighting term, and an adaptive
pseudo-labelling threshold that grows more conservative with distance from the labelled
region. Together with re-weighting these define the three-component framework reported as
a negative result in Section~\ref{sec:res-framework}.
 
\subsection{Data sources}
\label{sec:data}
 
\begin{figure}[H]
  \centering
  \includegraphics[width=0.98\linewidth]{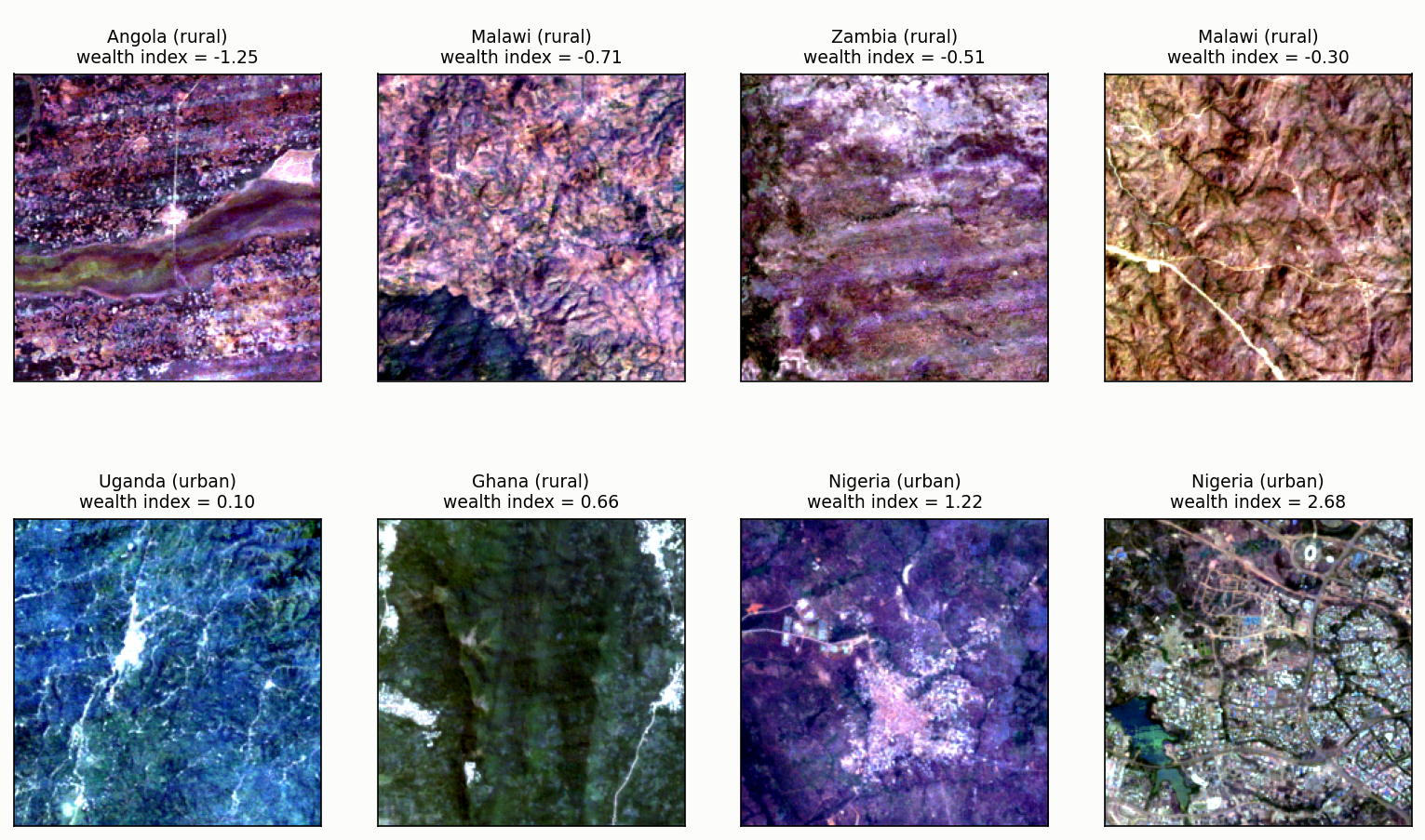}
  \caption{Eight PovertyMap-WILDS survey-cluster patches (false-colour composite of the red,
  green, and blue Landsat bands), spanning the observed asset-wealth range from lowest (top
  left) to highest (bottom right). Rural clusters at low wealth values show little built
  structure, while urban clusters at high wealth values show dense road networks and
  building footprints.}
  \label{fig:patches}
\end{figure}

PovertyMap-WILDS \citep{koh2021wilds} pairs satellite and nighttime-light imagery of
Demographic and Health Survey cluster locations with a continuous asset-wealth label and an
official country-level geographic split, making it a natural real-world analogue of the
synthetic mismatch axis. Figure~\ref{fig:patches} shows eight representative clusters
spanning the observed asset-wealth range, from sparsely developed rural terrain at the
lowest values to dense urban built environments at the highest. Each image patch is reduced
to a vector of per-band summary statistics, comprising the mean, standard deviation, and
the 10th, 50th, and 90th percentiles for each spectral and nighttime-light band, in the
spirit of the simple satellite-derived covariates used in the poverty-mapping literature
\citep{jean2016combining, rolf2021generalizable}, and the resulting 40-dimensional feature vector is
further reduced by principal component analysis to 5 components (92.8\% of variance
retained) so that the kernel-density-based
divergence estimators remain well posed at the sample sizes used here. The continuous label is
binarised at the population median so that the binary-classification methods and metrics
used throughout apply unchanged. Mismatch severity is operationalised discretely instead of
continuously: the labelled set at each severity level is drawn from a nested, shrinking
subset of the fold's official training countries, ordered alphabetically so that each level's
countries are a subset of the previous level's, while the unlabelled pool and in-region test
set span all training countries and the out-of-region test set is drawn from the fold's
official out-of-distribution countries, a stronger and more realistic notion of
out-of-region generalisation than a uniform-over-the-same-domain test set.
 
\begin{figure}[H]
  \centering
  \begin{subfigure}{0.32\linewidth}
    \includegraphics[width=\linewidth]{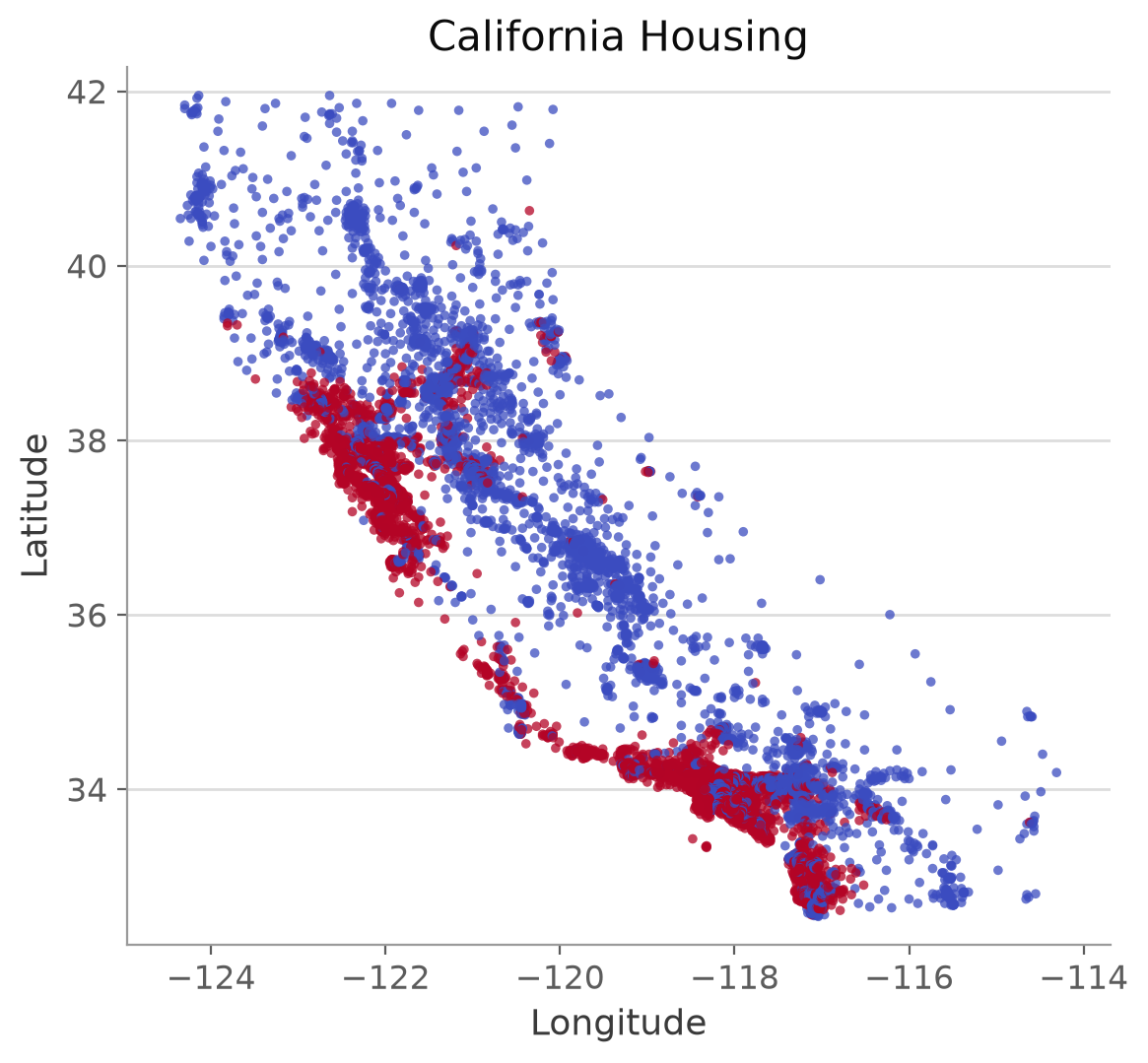}
    \caption{}
    \label{fig:realcova}
  \end{subfigure}
  \hfill
  \begin{subfigure}{0.32\linewidth}
    \includegraphics[width=\linewidth]{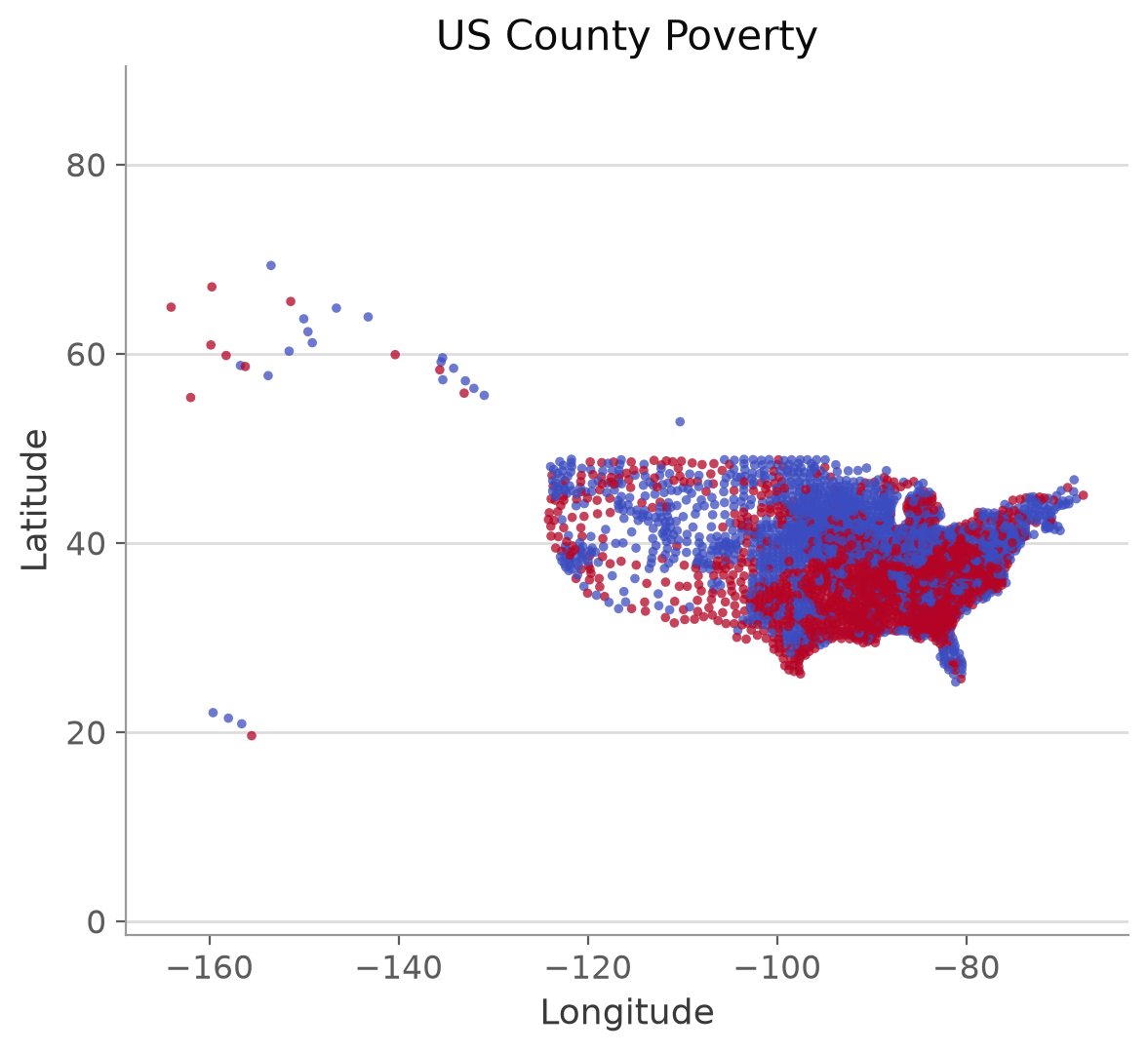}
    \caption{}
    \label{fig:realcovb}
  \end{subfigure}
  \hfill
  \begin{subfigure}{0.32\linewidth}
    \includegraphics[width=\linewidth]{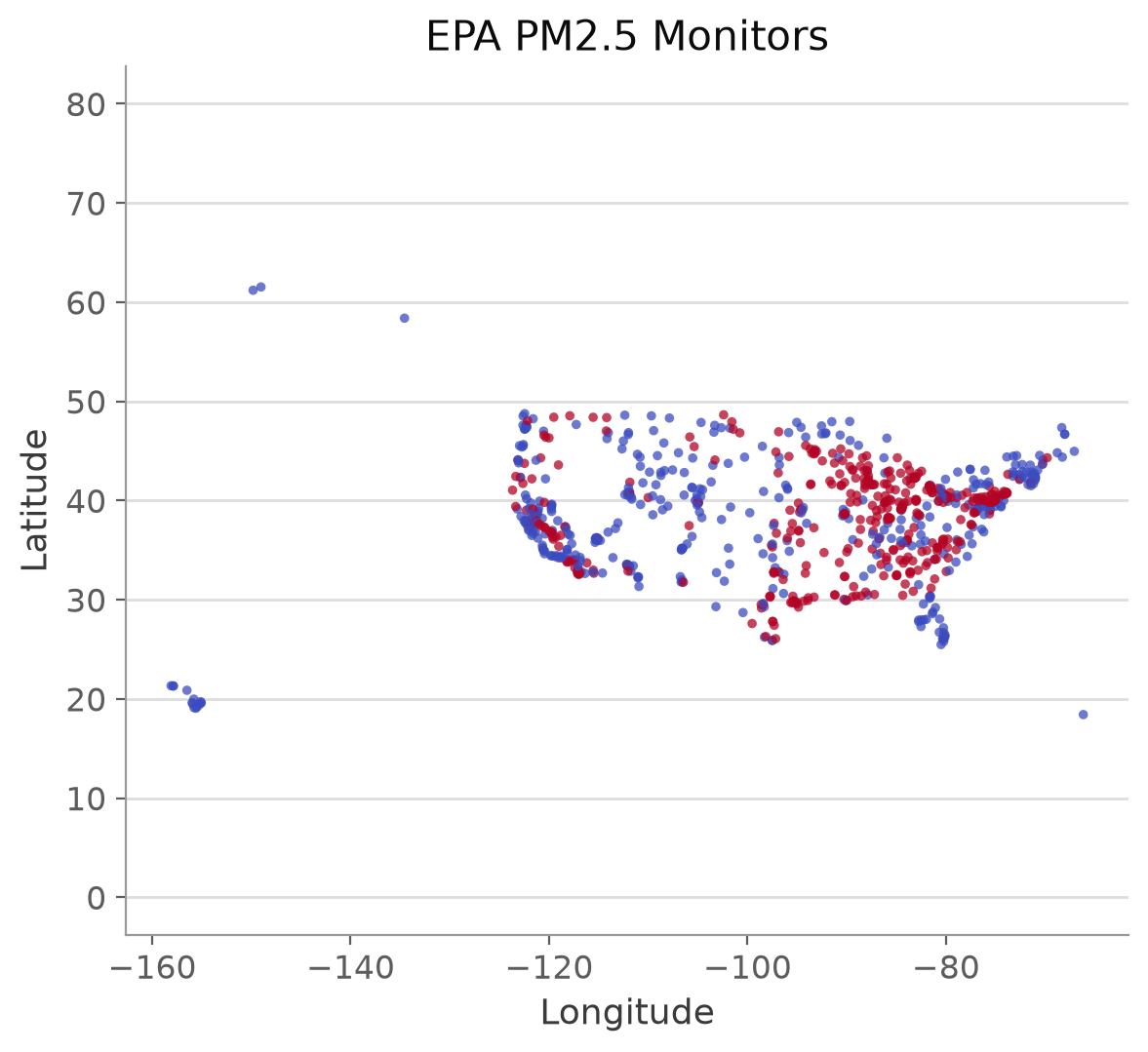}
    \caption{}
    \label{fig:realcovc}
  \end{subfigure}
  \caption{Geographic coverage and label distribution for the three real-world datasets:
  California Housing block groups (a), U.S.\ county poverty rates (b), and EPA
  PM$_{2.5}$ monitor sites (c), each coloured by whether the outcome is above or below its
  median.}
  \label{fig:realcov}
\end{figure}

Three further datasets instantiate the motivating application domains. For housing and the
built environment, California Housing \citep{pace1997sparse} provides 20,640 1990-census
block groups with median house value, income, age, room, and occupancy covariates and
centroid coordinates, each assigned to its nearest California county, giving 58 counties as
the geographic unit. For urban socio-economic modelling, the county-level data of the
Economic Research Service \citep{usda2024county} provide poverty, median household income,
unemployment, population, and educational-attainment estimates covering 3,130 counties across
50 states, joined to county centroid coordinates. For environmental and air-quality
monitoring, Air Quality System annual monitor summaries \citep{epa2024aqs} give 855
fine-particulate (PM$_{2.5}$) monitor sites paired with co-located ozone and coarse-particulate
concentrations and county population as land-use-regression-style covariates
\citep{hoek2008lur}. Figure~\ref{fig:realcov} shows each dataset's geographic coverage and
label distribution. For each, mismatch severity uses the same region-restriction design as
PovertyMap-WILDS, with county (housing) or state (socio-economic, air quality) as the
geographic unit; regions are split into training and out-of-distribution sets with a fixed
random seed, and training regions are ordered by ascending geographic distance from an anchor
region, defined as the largest training region by row count.

For each real dataset the binary label is a median split of the underlying continuous outcome
(house value, poverty rate, asset-wealth index, or PM$_{2.5}$ concentration), computed once
over the full dataset. Each dataset is a single-year snapshot:
1990 census block groups (Housing); 2023 county-level unemployment, income, population, and
educational-attainment tables merged by FIPS code, with complete-case deletion of any county
missing a required field (Socio-economic); 2023 Air Quality System annual monitor summaries,
with the ozone and coarse-particulate covariates imputed at their sample median for monitors
missing a co-located reading and counties missing a population estimate dropped (Air
quality); and the WILDS-provided satellite fold and asset-wealth label (PovertyMap-WILDS), with Table~\ref{tab:region-membership} stating how many regions play each role for every dataset.

\begin{table}[t]
  \centering
  \caption{Region membership by dataset. In-distribution regions supply the labelled
  candidates (a nested prefix, size set by $\alpha$), the unlabelled pool, and the in-region
  test set; out-of-distribution regions supply only the out-of-region test set. No
  validation split is used.}
  \label{tab:region-membership}
  \begin{tabular}{lccc}
    \toprule
    Dataset & Region unit & In-distribution & Out-of-distribution \\
    \midrule
    PovertyMap-WILDS & Country & 13 & 5 \\
    California Housing & County & 41 & 17 \\
    Socio-economic & State & 35 & 15 \\
    Air quality & State & 35 & 15 \\
    \bottomrule
  \end{tabular}
\end{table}

\subsection{Experimental protocol and changepoint analysis}
\label{sec:protocol}
 
The synthetic sweep crosses six mismatch levels ($\alpha \in \{0, 0.2, 0.4, 0.6, 0.8, 1.0\}$),
three autocorrelation length-scales, and two non-stationarity levels, for 324 runs; a subsequent re-run with twenty seeds per cell (2,160 runs) supports the sensitivity analysis. Unless otherwise noted, reported accuracies are
averaged over length-scale. Performance is evaluated by accuracy, F1, expected calibration error, and the spatial
generalisation gap $\Delta$. With predicted confidences binned into $B = 10$ equal-width bins
(the standard choice in \citealp{guo2017calibration}), expected calibration error is
$\mathrm{ECE} = \sum_{b=1}^{B} (|B_b|/N)\,\big|\mathrm{acc}(B_b) - \mathrm{conf}(B_b)\big|$
\citep{naeini2015obtaining, guo2017calibration}, a diagnostic that
distinguishes a model which is merely wrong from one which is wrong and confident
\citep{ovadia2019trust}.
 
To locate the breakdown formally instead of reading it off a curve, a two-segment
piecewise-linear model is fitted to out-of-region accuracy, and separately to $\Delta$, as a
function of $\alpha$ and separately per method \citep{muggeo2003segmented},
\begin{equation}
A(\alpha) = \beta_0 + \beta_1 \alpha + \beta_2 (\alpha - \alpha^{\star})_{+} + \varepsilon,
\label{eq:segmented}
\end{equation}
where $(\cdot)_+$ denotes the positive part and $\alpha^{\star}$ the breakpoint. The breakpoint
is estimated on the twenty-seed re-run and given a percentile bootstrap confidence interval by
resampling seeds \citep{efron1993bootstrap}; since repeated runs and observations within a
region share spatial units, this seed-level resampling is complemented by the region-split and
country-ordering sensitivity check in the spirit of
resampling methods for dependent data more broadly \citep{lahiri2003resampling}.
\newpage
\section{Results}
 
\subsection{Effect of marginal distribution mismatch}
\label{sec:res-mismatch}

\begin{figure}[H]
  \centering
  \begin{subfigure}{0.49\linewidth}
    \includegraphics[width=\linewidth]{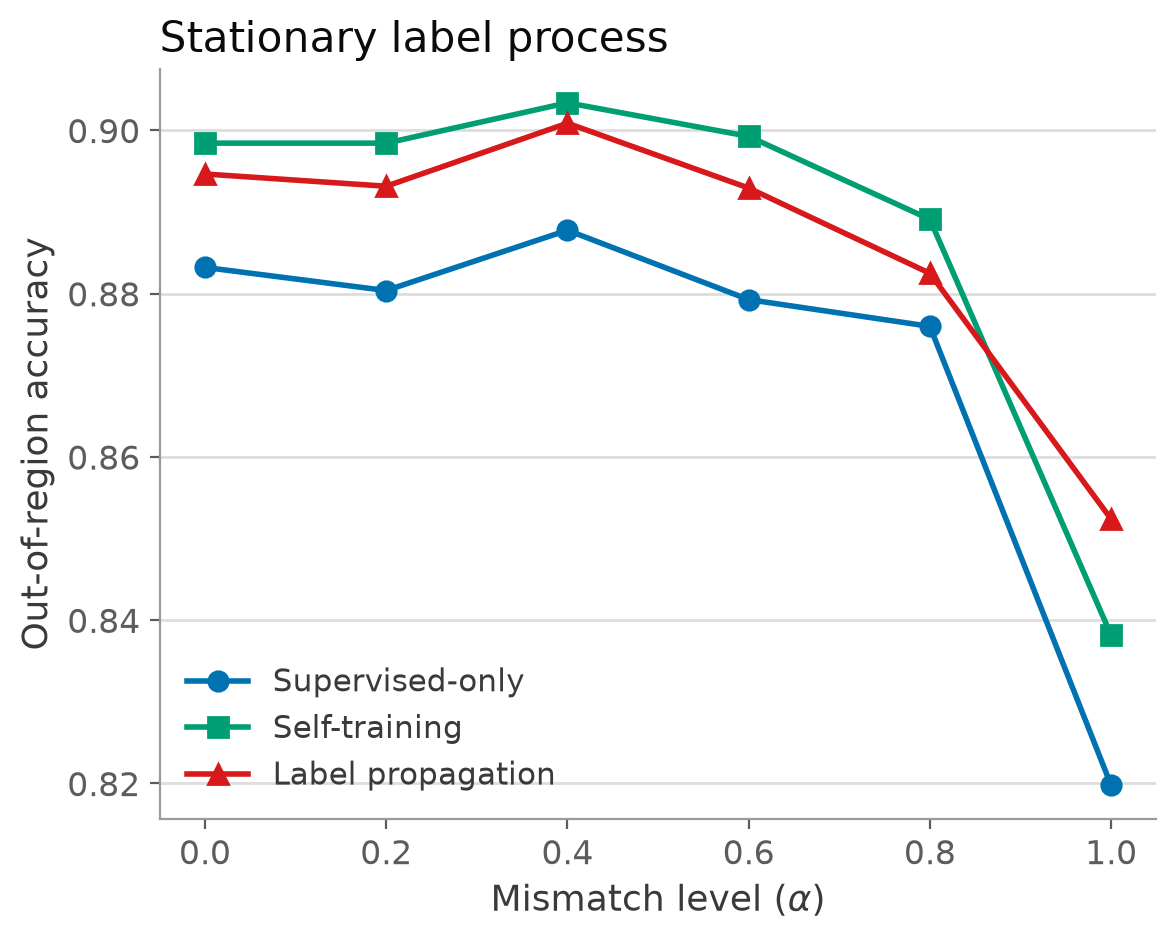}
    \caption{}
    \label{fig:acc-mismatch-a}
  \end{subfigure}
  \hfill
  \begin{subfigure}{0.49\linewidth}
    \includegraphics[width=\linewidth]{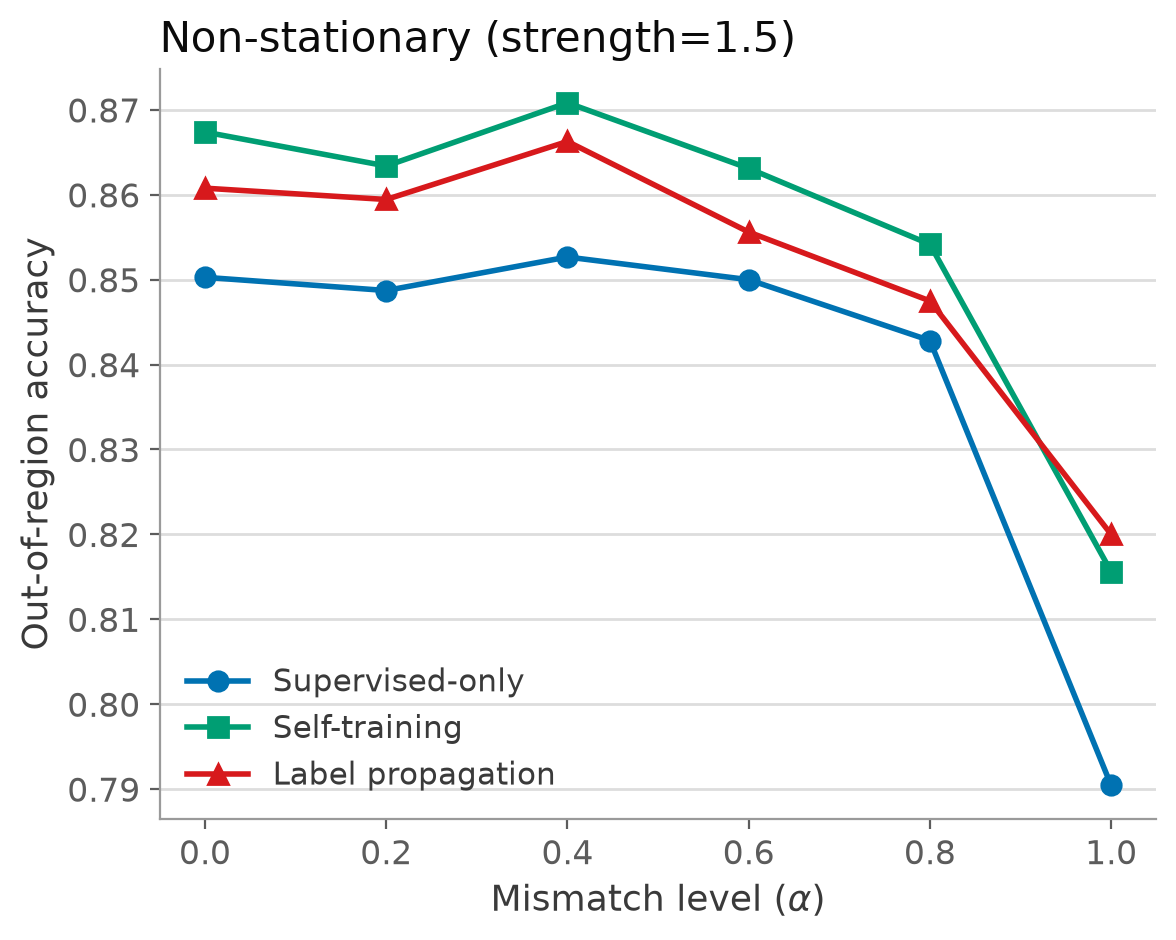}
    \caption{}
    \label{fig:acc-mismatch-b}
  \end{subfigure}
  \caption{Out-of-region accuracy as a function of marginal distribution mismatch $\alpha$, by
  method, for (a) a stationary label process and (b) under non-stationarity. Accuracy
  remains relatively stable for $\alpha \le 0.6$ before declining sharply at higher mismatch
  levels for all three methods.}
  \label{fig:acc-mismatch}
\end{figure}

\begin{figure}[H]
  \centering
  \includegraphics[width=0.7\linewidth]{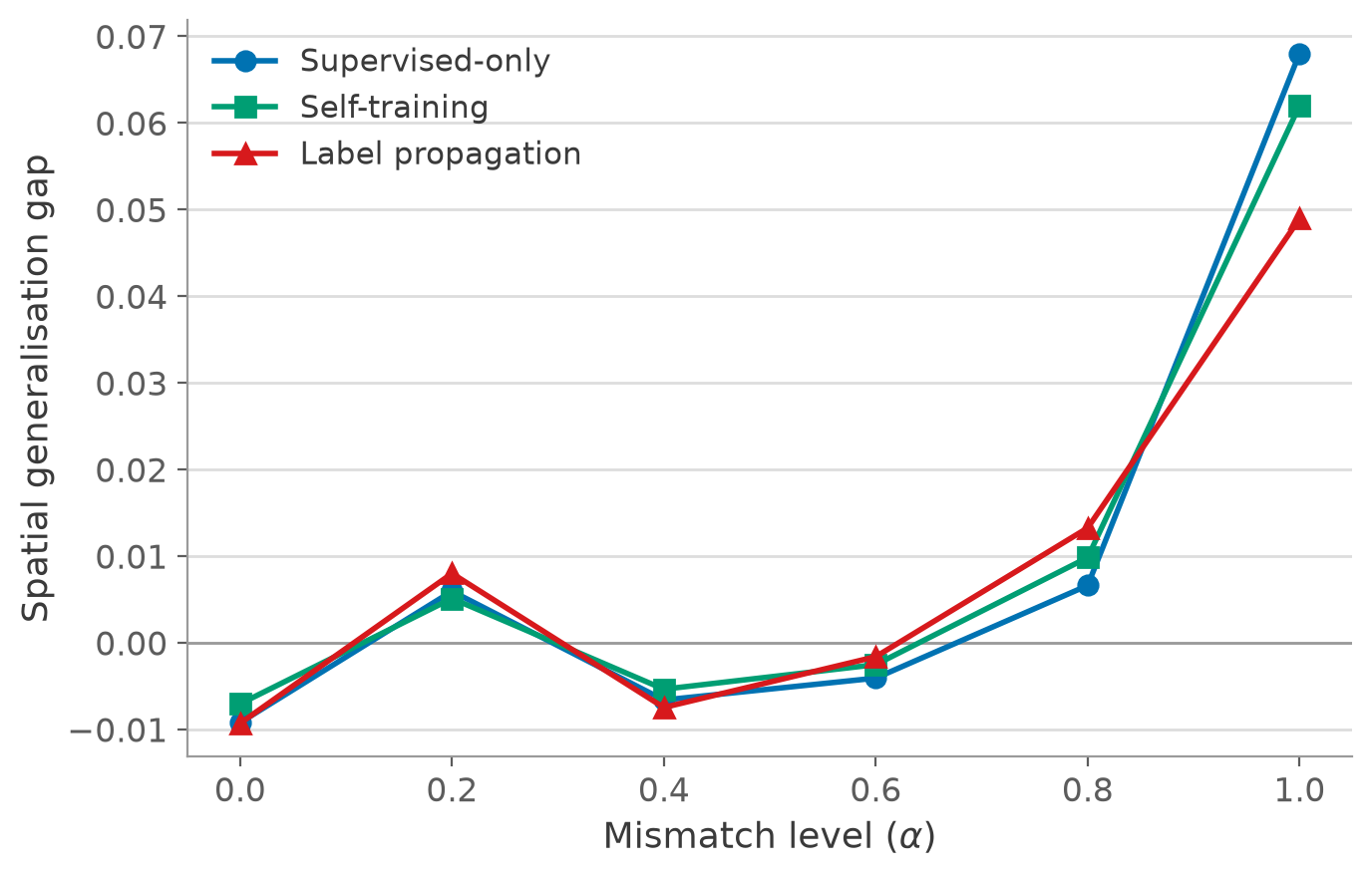}
  \caption{Spatial generalisation gap (in-region accuracy minus out-of-region accuracy) as a
  function of marginal distribution mismatch $\alpha$. The gap remains close to zero for
  $\alpha \le 0.6$ before widening substantially at $\alpha = 1$, indicating increasing
  overfitting to the labelled region.}
  \label{fig:gap-mismatch}
\end{figure}

Out-of-region test accuracy declined for all three methods as marginal mismatch increased from
$\alpha = 0$ to $\alpha = 1$ (Table~\ref{tab:t1}). Accuracy drops ranged from 4.2 percentage
points for label propagation to 6.2 for supervised-only. Although supervised-only showed the
largest point-estimate decline, the 95\% seed-cluster bootstrap confidence intervals for the
three methods overlap. All three methods are fit on identical seed-level draws (the same
labelled, unlabelled, and test split at each $\alpha$, length-scale, and non-stationarity
combination), hence, this marginal overlap is not itself a test of whether their degradation differs.
A paired seed-cluster bootstrap on the between-method difference in drop, resampling whole seeds
and computing each pair's difference on the same resample, shows that label propagation's drop is
reliably smaller than both supervised-only's (2.0 percentage points, 95\% CI $[0.6,\,3.6]$) and
self-training's (1.5 percentage points, 95\% CI $[0.1,\,3.0]$), while supervised-only and
self-training remain statistically indistinguishable from each other (0.6 percentage points, 95\%
CI $[-1.0,\,2.0]$). This drop comparison, however, does not by itself show that the
unlabelled data are responsible for it: labelled coverage also shrinks as $\alpha$
increases, so supervised-only's decline reflects fewer labels, not unlabelled-data harm.
Table~\ref{tab:t1}'s last two columns instead compare each SSL method against
supervised-only fit on the identical draw at a fixed $\alpha$; both beat supervised-only at
every mismatch level, with CIs excluding zero throughout (Supplementary Table~S6). The
unlabelled data are never harmful relative to not using them at all, in this sweep. The
relationship between mismatch and accuracy was not linear. Performance
remained relatively stable for $\alpha \le 0.6$ before declining sharply as mismatch increased
further (Figure~\ref{fig:acc-mismatch}). The spatial generalisation gap showed the same pattern,
remaining close to zero at lower mismatch levels before widening to 0.05--0.07 at
$\alpha = 1$ (Figure~\ref{fig:gap-mismatch}). This indicates that severe mismatch causes models
to generalise increasingly poorly beyond the regions from which labelled data were collected.
These results support H1 and suggest that the effect of marginal mismatch is best
characterised as a threshold-like breakdown instead of a gradual decline in performance, with
label propagation degrading measurably more gracefully than the other two baselines.

\begin{table}[t]
  \centering
  \caption{Out-of-region accuracy at minimal ($\alpha=0$) versus maximal ($\alpha=1$) marginal
  distribution mismatch, averaged over length-scale, non-stationarity level, and seed, with
  twenty seeds per cell. Drop is reported in percentage points (pp) with its 95\% seed-cluster
  bootstrap confidence interval. The last two columns are a paired comparison against
  supervised-only, fit on the same seed-level draws, at each of the two endpoints
  (Supplementary Table~S6 reports all six mismatch levels).}
  \label{tab:t1}
  \begin{tabular}{lccccc}
    \toprule
    Method & $\alpha=0$ & $\alpha=1$ & Drop (pp) & \multicolumn{2}{c}{$\Delta$ vs Sup.\ (pp)} \\
     & & & 95\% CI & $\alpha=0$ & $\alpha=1$ \\
    \midrule
    Supervised-only    & 0.867 & 0.805 & 6.2 $[4.5,\,7.8]$    & --                  & -- \\
    Self-training      & 0.883 & 0.827 & 5.6 $[3.9,\,7.4]$    & $+1.6\,[1.2,\,2.0]$ & $+2.2\,[0.5,\,3.4]$ \\
    Label propagation  & 0.878 & 0.836 & 4.2 $[2.5,\,5.8]$    & $+1.1\,[0.8,\,1.4]$ & $+3.1\,[1.6,\,4.4]$ \\
    \bottomrule
  \end{tabular}
\end{table}



\subsection{Cluster and manifold sensitivity}
\label{sec:res-sensitivity-mech}

The synthetic experiments provide only partial support for H2. Self-training is often expected to be more sensitive to distribution mismatch, given its reliance on pseudo-labels, yet it does not exhibit the largest performance drop in this experiment: its estimated decline (5.6 pp) is slightly smaller than the supervised-only baseline's (6.2 pp), and the paired seed-cluster bootstrap of Section~\ref{sec:res-mismatch} confirms the two are not statistically distinguishable. Label propagation shows the smallest decline (4.2 pp), and the same bootstrap shows this is a real effect rather than a favourable point estimate: its drop is reliably smaller than both baselines'. This indicates that it is more robust to marginal mismatch than the other two methods in the synthetic setting. The real-data evaluation on PovertyMap-WILDS (Section~\ref{sec:res-wilds}) shows self-training with the largest point-estimate reduction in out-of-region accuracy among the three baselines.


\subsection{Spatial non-stationarity in isolation}
\label{sec:res-nonstat}

At $\alpha = 0$, where the labelled and unlabelled data are drawn from the same spatial
distribution, increasing the non-stationarity strength from 0 to 1.5 reduced accuracy by
approximately 3.3 percentage points across methods in the twenty-seed experiments, with
supervised-only accuracy decreasing from 0.883 to 0.850. As this reduction occurs in the
absence of marginal mismatch, it shows that spatial non-stationarity, a form of concept shift
defined in Equation~\eqref{eq:concshift}, degrades performance independently of covariate shift
(Equation~\eqref{eq:covshift}). The magnitude of this effect is comparable to that of severe marginal mismatch, which reduces
accuracy by approximately 4--6 percentage points across methods. The non-stationarity effect is
close to additive, not interacting with mismatch severity: accuracy at
non-stationarity~$=1.5$ is 2.8--3.4 percentage points below non-stationarity~$=0$ at every
value of $\alpha$ tested (3.3, 3.4, 3.4, 3.4, 3.4, and 2.8 points at $\alpha = 0, 0.2, \ldots,
1.0$ respectively), instead of the gap widening or narrowing systematically with severity.
Together, these results show that marginal mismatch and spatial non-stationarity contribute
substantially and largely independently to performance degradation, consistent with them being
distinct mechanisms instead of one driving the other.

\subsection{Divergence metrics as proxies for performance loss}
\label{sec:res-div}

\begin{figure}[H]
  \centering
  \begin{subfigure}{0.5\linewidth}
    \includegraphics[width=\linewidth]{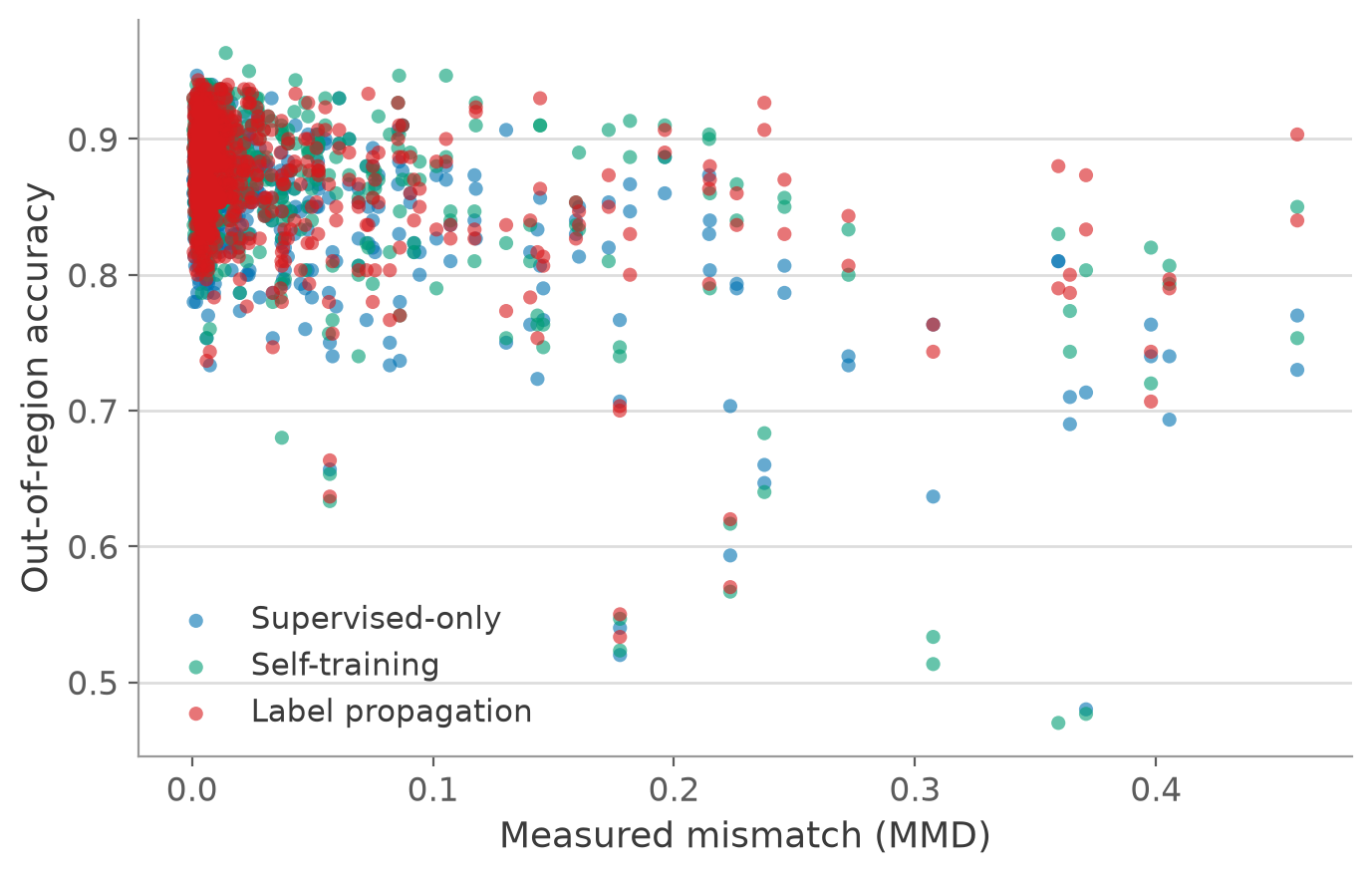}
    \caption{}
    \label{fig:mmd-scatter}
  \end{subfigure}\hfill
  \begin{subfigure}{0.5\linewidth}
    \includegraphics[width=\linewidth]{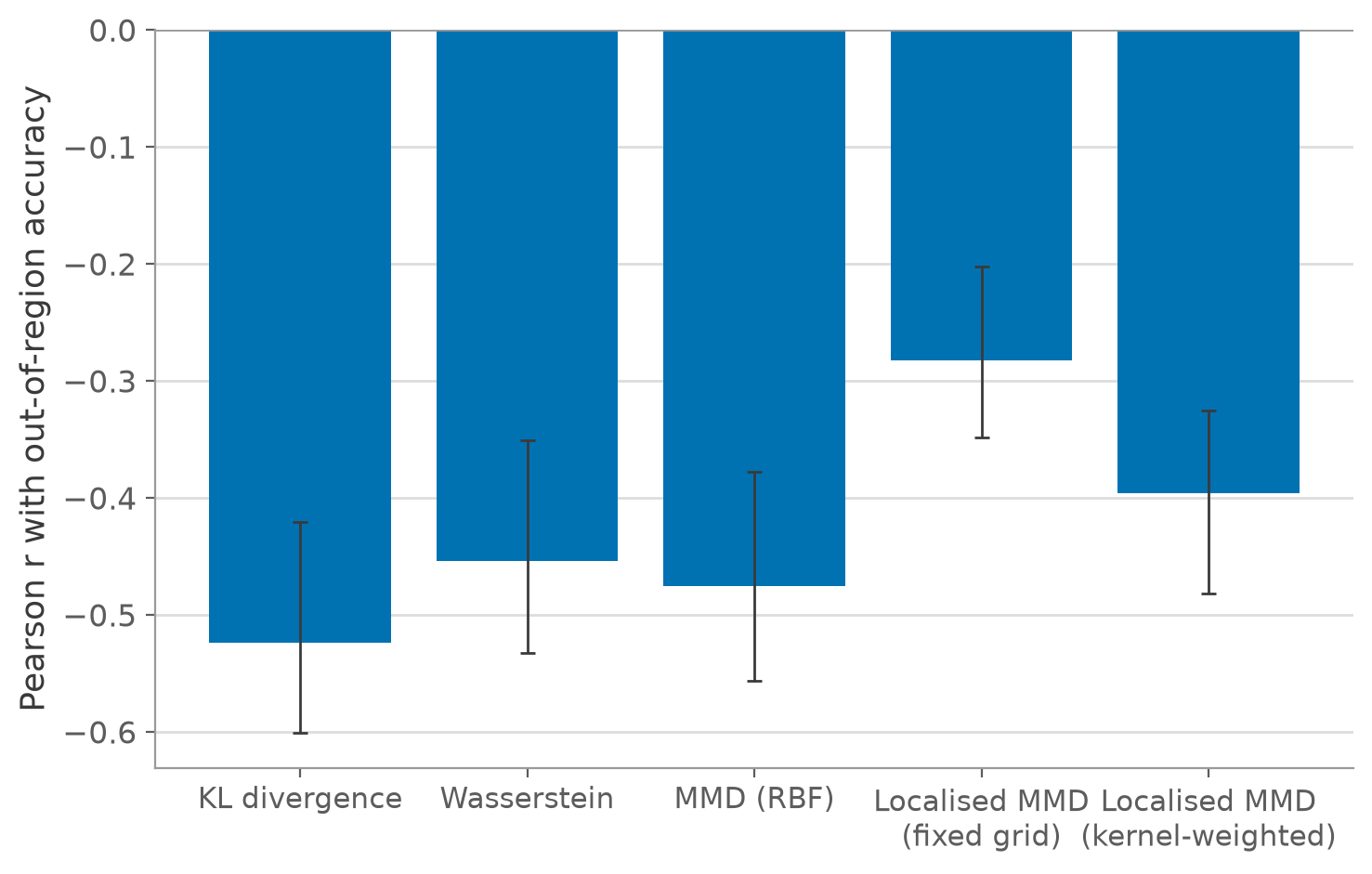}
    \caption{}
    \label{fig:corr-bars}
  \end{subfigure}
  \caption{(a) Out-of-region accuracy against measured mismatch across all 2,160 runs; global divergence metrics track accuracy more reliably than the original fixed-grid localised variant. (b) Pooled Pearson correlation between each divergence metric and out-of-region accuracy; correlations are negative throughout, since greater measured mismatch accompanies lower accuracy. The kernel-weighted localised estimator raises the magnitude from $r=-0.28$ to $r=-0.40$, closing part but not most of the gap to the global metrics.}
  \label{fig:div}
\end{figure}

To evaluate whether the divergence metrics introduced in
Section~\ref{sec:divergence} provide useful indicators of performance degradation, we computed the pooled correlation between each metric and out-of-region accuracy across all 2,160
synthetic runs (Table~\ref{tab:t2}, Figures~\ref{fig:mmd-scatter}
and~\ref{fig:corr-bars}). Confidence intervals were estimated using seed-cluster bootstrap resampling. The global divergence metrics showed the strongest relationship with performance. Based on
Pearson correlation, Kullback--Leibler divergence, maximum mean discrepancy, and the
Wasserstein distance achieved correlations of $r=-0.52$, $-0.48$, and $-0.45$,
respectively, with all 95\% confidence intervals excluding zero. Kullback--Leibler
divergence remained similarly informative when the analysis was restricted to the most
severe mismatch level ($\alpha=1.0$), with a correlation of $r=-0.50$. Spearman's rank correlations showed the same overall ordering of the metrics, although with
smaller magnitudes (Table~\ref{tab:t2}). Kullback--Leibler divergence again exhibited the
strongest association with performance ($\rho=-0.31$), followed by the kernel-weighted
local MMD ($\rho=-0.25$), while the fixed-grid local estimator remained the weakest
($\rho=-0.11$). The consistent ranking across both Pearson and Spearman correlations
indicates that the relative usefulness of the divergence metrics is robust to whether the
relationship is assessed linearly or by monotonic association. The localised divergence estimates were less strongly correlated with performance than the
global metrics. The original fixed-grid estimator achieved a Pearson correlation of only
$r=-0.28$, whereas the proposed kernel-weighted estimator increased this to $r=-0.40$.
Although the improvement does not fully match the performance of the global metrics, it
substantially strengthens the local estimate while retaining the ability to identify where
mismatch is concentrated.

The pooled correlations above are computed across all six mismatch levels together, so part
of the signal could simply reflect the shared trend that both divergence and inaccuracy rise
with $\alpha$, instead of the metric distinguishing higher- from lower-risk runs at a fixed
severity. We tested this directly in two ways. First, a partial correlation that residualises
both accuracy and each divergence metric on $\alpha$, length-scale, and non-stationarity
strength before correlating them retains a moderate signal for the global metrics
(Kullback--Leibler $r=-0.48$, maximum mean discrepancy $r=-0.40$, Wasserstein $r=-0.37$, all
$p<10^{-70}$) but nearly halves it for the kernel-weighted local estimator ($r=-0.20$,
$p<10^{-20}$). Second, recomputing the Pearson correlation separately within each of the six
$\alpha$ levels shows that the local estimator's correlation with accuracy is not
distinguishable from zero at any single level (all $|r|<0.16$, most $p>0.1$), whereas the
plain (non-localised) MMD retains a usable within-level signal only at the two most severe
levels ($\alpha=0.8$: $r=-0.47$; $\alpha=1.0$: $r=-0.39$) and is weak or inconsistent in sign
below that. The kernel-weighted local estimator therefore functions mainly as an indicator of
which severity regime a run belongs to, not as a diagnostic that predicts which individual
run will underperform within a fixed severity level; using it as a per-deployment reliability
signal, as opposed to a coarse indicator of overall mismatch severity, is not supported by
this analysis.

\begin{table}[t]
  \centering
  \caption{Pooled Pearson ($r$) and Spearman ($\rho$) correlation between distribution-divergence
  metrics and out-of-region accuracy across all 2,160 synthetic runs (twenty seeds per cell), with
  95\% seed-cluster bootstrap confidence intervals accounting for the seed/length-scale/
  non-stationarity nesting.}
  \label{tab:t2}
  \begin{tabular}{lcccc}
    \toprule
     & \multicolumn{2}{c}{Pearson} & \multicolumn{2}{c}{Spearman} \\
    \cmidrule(lr){2-3} \cmidrule(lr){4-5}
    Metric & $r$ & 95\% CI & $\rho$ & 95\% CI \\
    \midrule
    Kullback--Leibler divergence (KDE)     & $-0.52$ & $[-0.60,\,-0.42]$ & $-0.31$ & $[-0.39,\,-0.23]$ \\
    Maximum mean discrepancy (RBF kernel)  & $-0.48$ & $[-0.56,\,-0.38]$ & $-0.22$ & $[-0.32,\,-0.12]$ \\
    Wasserstein distance (marginal)        & $-0.45$ & $[-0.53,\,-0.35]$ & $-0.22$ & $[-0.31,\,-0.13]$ \\
    Localised MMD ($5\times5$ fixed grid)  & $-0.28$ & $[-0.35,\,-0.20]$ & $-0.11$ & $[-0.18,\,-0.04]$ \\
    Localised MMD (kernel-weighted)        & $-0.40$ & $[-0.48,\,-0.33]$ & $-0.25$ & $[-0.32,\,-0.16]$ \\
    \bottomrule
  \end{tabular}
\end{table}
 
\subsection{Effect of mismatch on calibration and accuracy}
\label{sec:res-calib}

Out-of-region F1 broadly mirrors the accuracy results in Table~\ref{tab:t1}. Self-training
declines from 0.885 to 0.807, supervised-only from 0.870 to 0.794, and label propagation from
0.880 to 0.828. As with accuracy, the reductions for self-training and supervised-only (0.078
and 0.076 points, respectively) are too similar to establish a clear difference in robustness.

Calibration exhibits a more consistent pattern. Out-of-region expected calibration error
roughly doubles for all three methods as mismatch increases from $\alpha = 0$ to
$\alpha = 1$, whereas in-region calibration error remains stable or improves slightly
(Table~\ref{tab:t3}). This indicates that severe mismatch affects not only predictive accuracy
but also the reliability of model confidence outside the labelled region.

Among the three methods, self-training shows the largest increase in out-of-region calibration
error (0.062 points), compared with 0.041 for supervised-only and 0.038 for label propagation.
Although the accuracy differences between methods are small, calibration reveals greater
variation in their robustness under severe mismatch.

\begin{table}[t]
  \centering
  \caption{Expected calibration error, in-region and out-of-region, at minimal ($\alpha=0$) versus
  maximal ($\alpha=1$) mismatch, averaged over length-scale, non-stationarity level, and seed,
  twenty seeds per cell. Lower values indicate better calibration.}
  \label{tab:t3}
  \begin{tabular}{lcccc}
    \toprule
    & \multicolumn{2}{c}{ECE (in-region)} & \multicolumn{2}{c}{ECE (out-of-region)} \\
    \cmidrule(lr){2-3}\cmidrule(lr){4-5}
    Method & $\alpha=0$ & $\alpha=1$ & $\alpha=0$ & $\alpha=1$ \\
    \midrule
    Self-training     & 0.082 & 0.070 & 0.075 & 0.137 \\
    Supervised-only   & 0.044 & 0.046 & 0.039 & 0.080 \\
    Label propagation & 0.038 & 0.035 & 0.035 & 0.073 \\
    \bottomrule
  \end{tabular}
\end{table}
 
\subsection{Benchmark validation on PovertyMap-WILDS}
\label{sec:res-wilds}

\begin{figure}[H]
  \centering
  \begin{subfigure}{0.49\linewidth}
    \includegraphics[width=\linewidth]{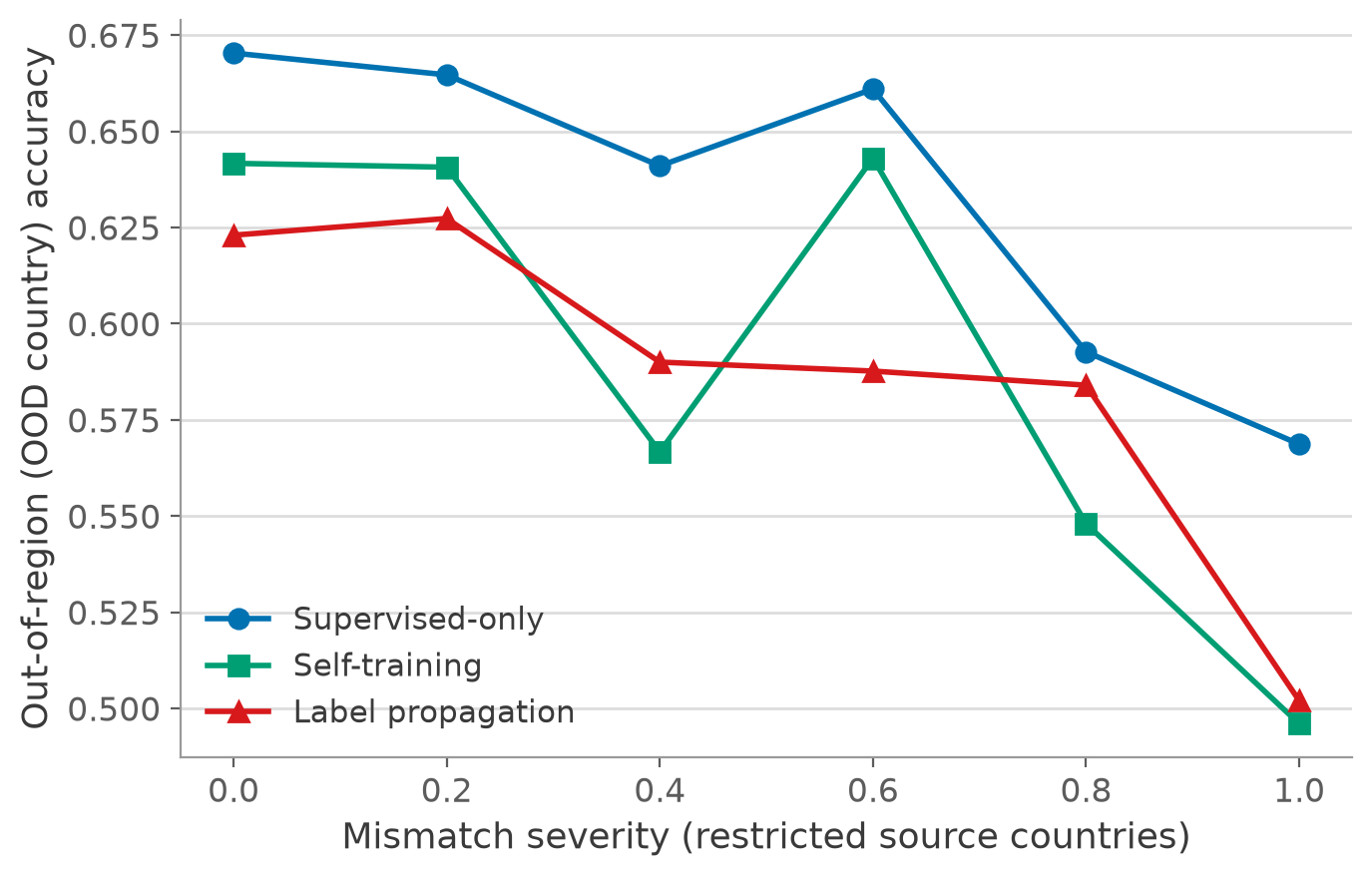}
    \caption{}
    \label{fig:wilds-out}
  \end{subfigure}\hfill
  \begin{subfigure}{0.49\linewidth}
    \includegraphics[width=\linewidth]{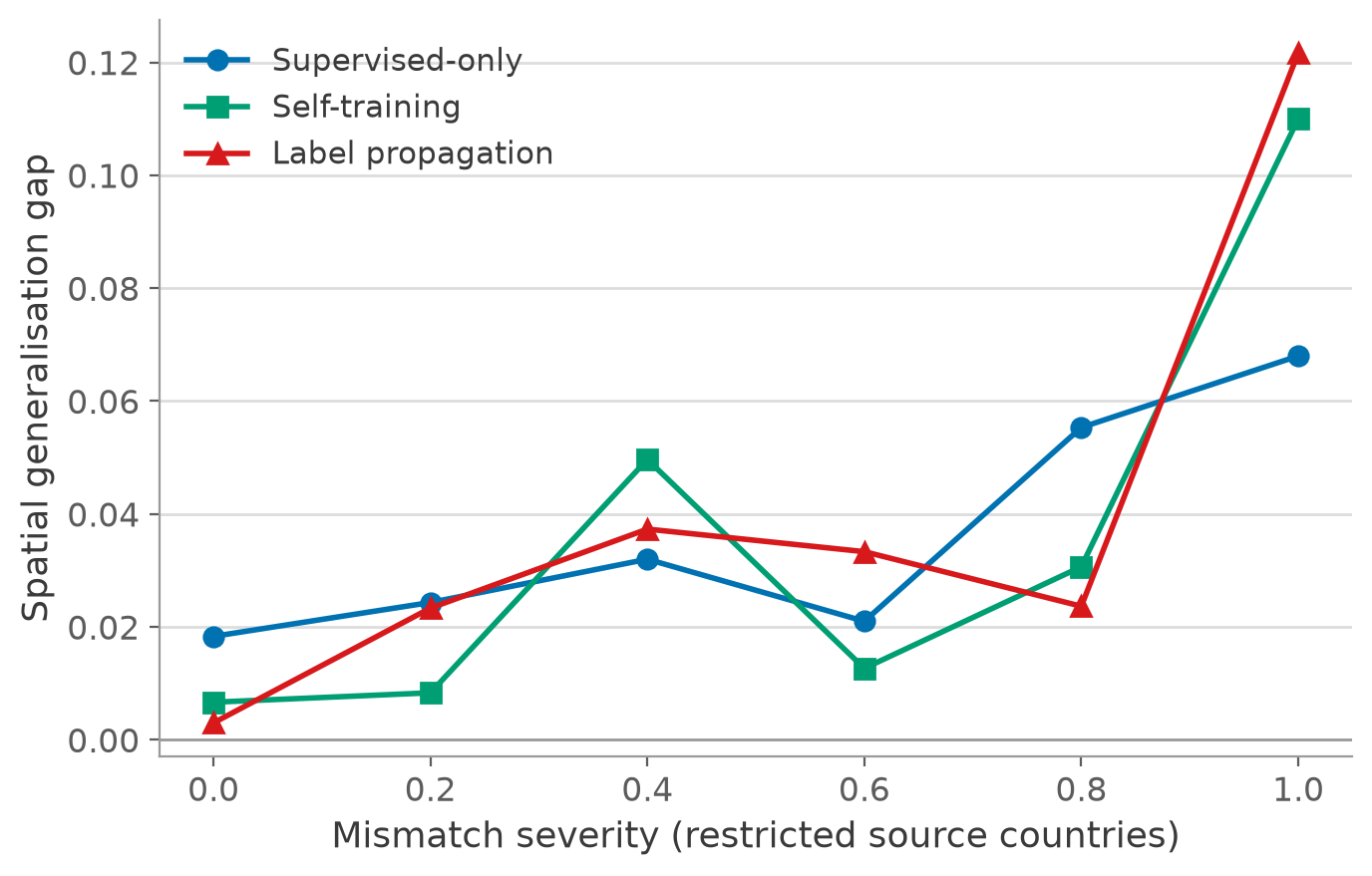}
    \caption{}
    \label{fig:wilds-gap}
  \end{subfigure}
  \caption{PovertyMap-WILDS, by method: (a) out-of-region accuracy and (b) spatial generalisation gap,
  each as a function of mismatch severity $\alpha \in \{0,0.2,\ldots,1.0\}$, corresponding to labelled
  sets drawn from 13, 11, 8, 6, 3, and 1 of the training countries respectively. All three methods are
  comparatively stable while several countries remain available, then decline once the labelled set is
  restricted to one to three countries, most sharply for self-training.}
  \label{fig:wilds}
\end{figure}

To assess whether the synthetic findings generalise to a real spatial benchmark, we repeated the
controlled-mismatch experiment on PovertyMap-WILDS \citep{koh2021wilds} using the same three
baseline methods across six mismatch-severity levels, where the labelled data were drawn from
13, 11, 8, 6, 3, and finally 1 of the training countries. The evaluation comprised 360 runs
using twenty random seeds (Figures~\ref{fig:wilds-out}
and~\ref{fig:wilds-gap}).

The threshold shape observed in the synthetic experiments is reproduced: all three methods remain comparatively stable while several countries remain available, then decline once the labelled set is restricted to one to three countries (Figure~\ref{fig:wilds}). The method ordering, however, differs. Self-training shows the largest point-estimate reduction in out-of-region accuracy, dropping
by 14.6 percentage points (95\% CI $[9.7,\,21.4]$), compared with 12.1 for label propagation
and 10.2 for the supervised-only baseline (Table~\ref{tab:t4}). As on the synthetic data, we
checked this with a paired seed-cluster bootstrap on the same seed-level draws instead of
relying on the overlap of these marginal intervals: none of the three pairwise differences is
distinguishable from zero (self-training vs.\ supervised-only, $95\%$ CI $[-1,\,11]$ pp;
self-training vs.\ label propagation, $[-2,\,8]$ pp; label propagation vs.\ supervised-only,
$[-1,\,5]$ pp). PovertyMap-WILDS therefore does not, on its own, establish that self-training
is the most sensitive of the three methods; it is directionally consistent with that reading
but not statistically conclusive at this sample size.

One aspect of the synthetic results does not fully replicate. Whereas label propagation showed
the smallest performance decline in the synthetic experiments, supervised-only performs best on
PovertyMap-WILDS, with label propagation lying between the two methods. However, the confidence
intervals for these two methods overlap, so this difference should be interpreted cautiously,
not as a clear reversal of their relative robustness.

\begin{table}[t]
  \centering
  \caption{PovertyMap-WILDS: out-of-region (out-of-distribution country) accuracy at minimal (13
  training countries) versus maximal (1 training country) mismatch severity, averaged over seed,
  twenty seeds per level. Drop is reported with its 95\% seed-cluster bootstrap confidence interval.}
  \label{tab:t4}
  \begin{tabular}{lcccc}
    \toprule
    Method & 13 countries & 1 country & Drop (pp) & 95\% CI \\
    \midrule
    Self-training     & 0.642 & 0.496 & 14.6 & $[9.7,\,21.4]$ \\
    Label propagation & 0.623 & 0.502 & 12.1 & $[10.1,\,14.1]$ \\
    Supervised-only   & 0.670 & 0.569 & 10.2 & $[6.9,\,13.7]$ \\
    \bottomrule
  \end{tabular}
\end{table}
 
\subsection{Validation on real datasets}
\label{sec:res-real}

\begin{figure}[t]
  \centering
  \begin{subfigure}{0.32\linewidth}
    \includegraphics[width=\linewidth]{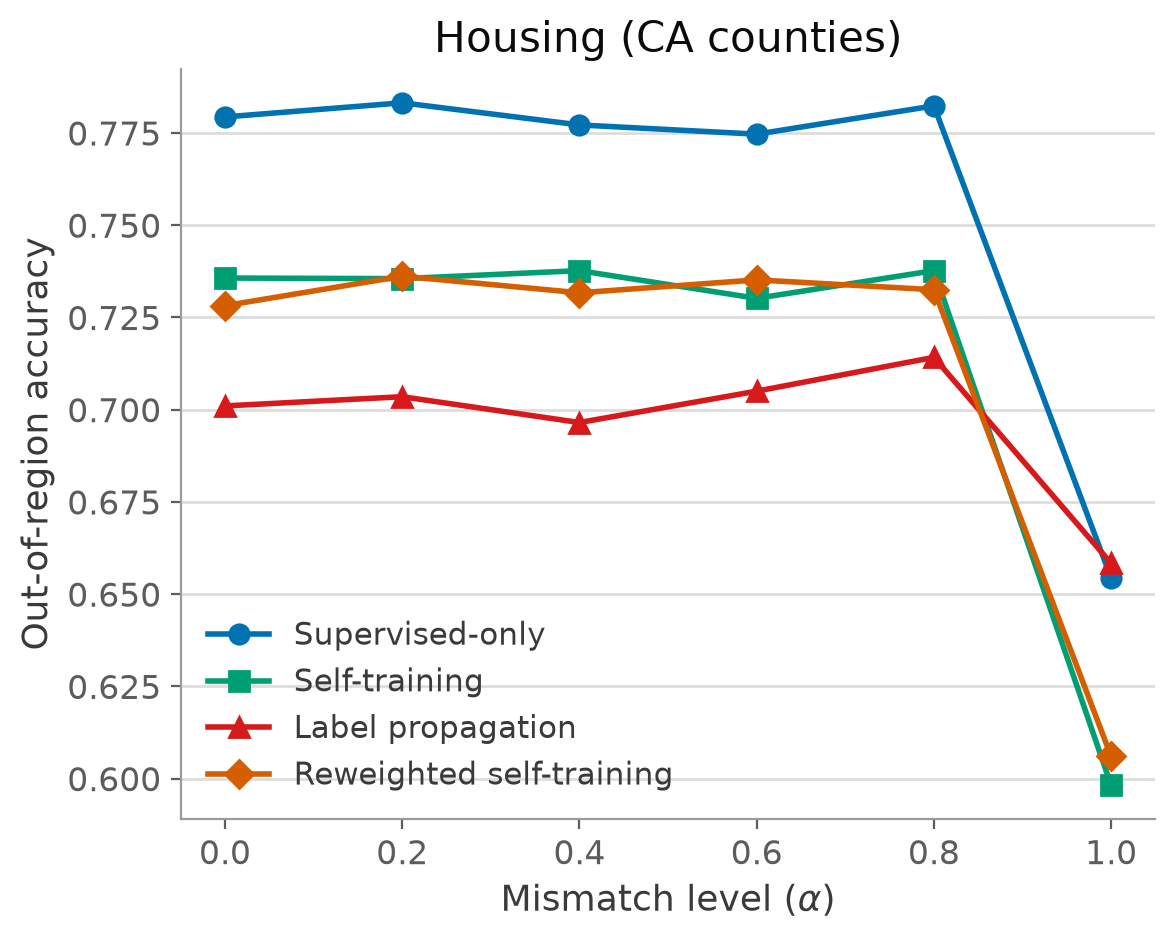}
    \caption{}
    \label{fig:real-a}
  \end{subfigure}
  \hfill
  \begin{subfigure}{0.32\linewidth}
    \includegraphics[width=\linewidth]{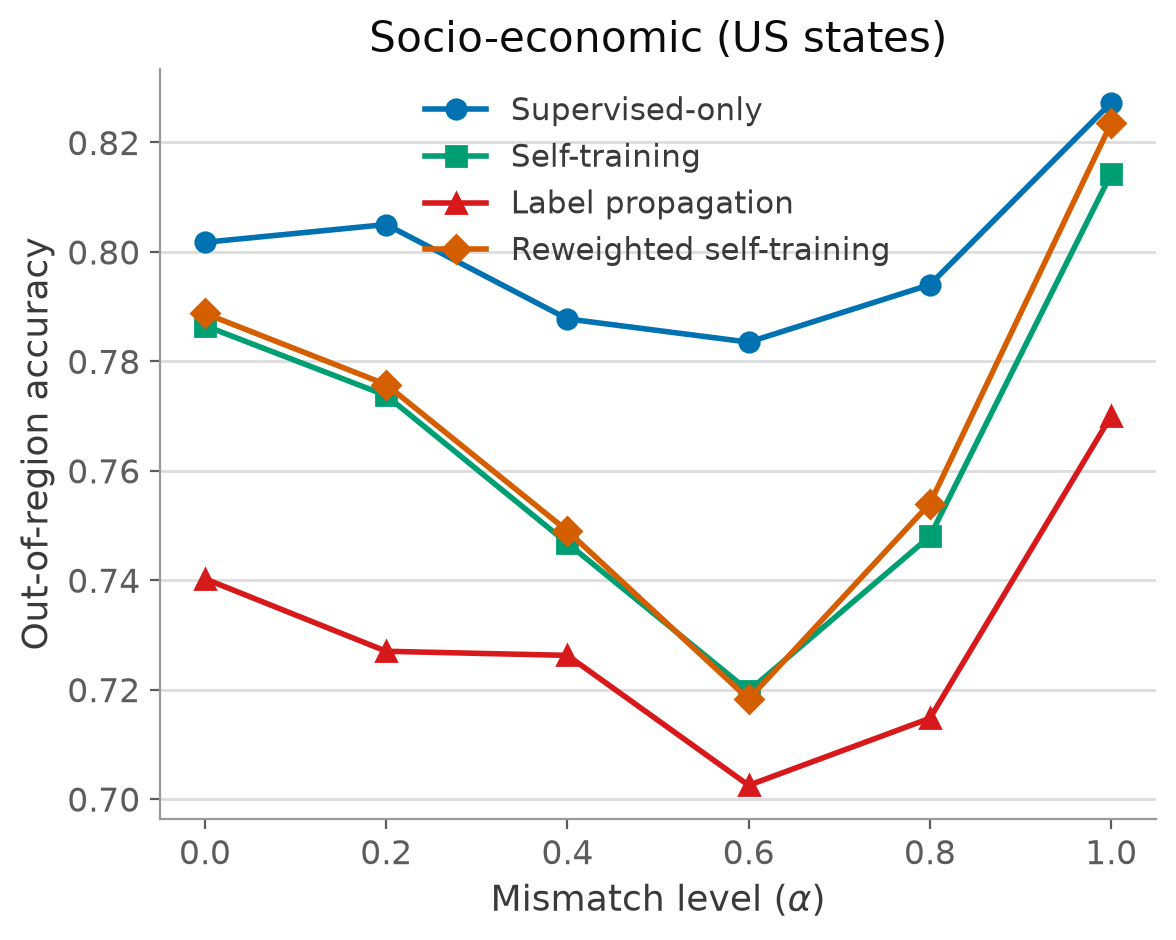}
    \caption{}
    \label{fig:real-b}
  \end{subfigure}
  \hfill
  \begin{subfigure}{0.32\linewidth}
    \includegraphics[width=\linewidth]{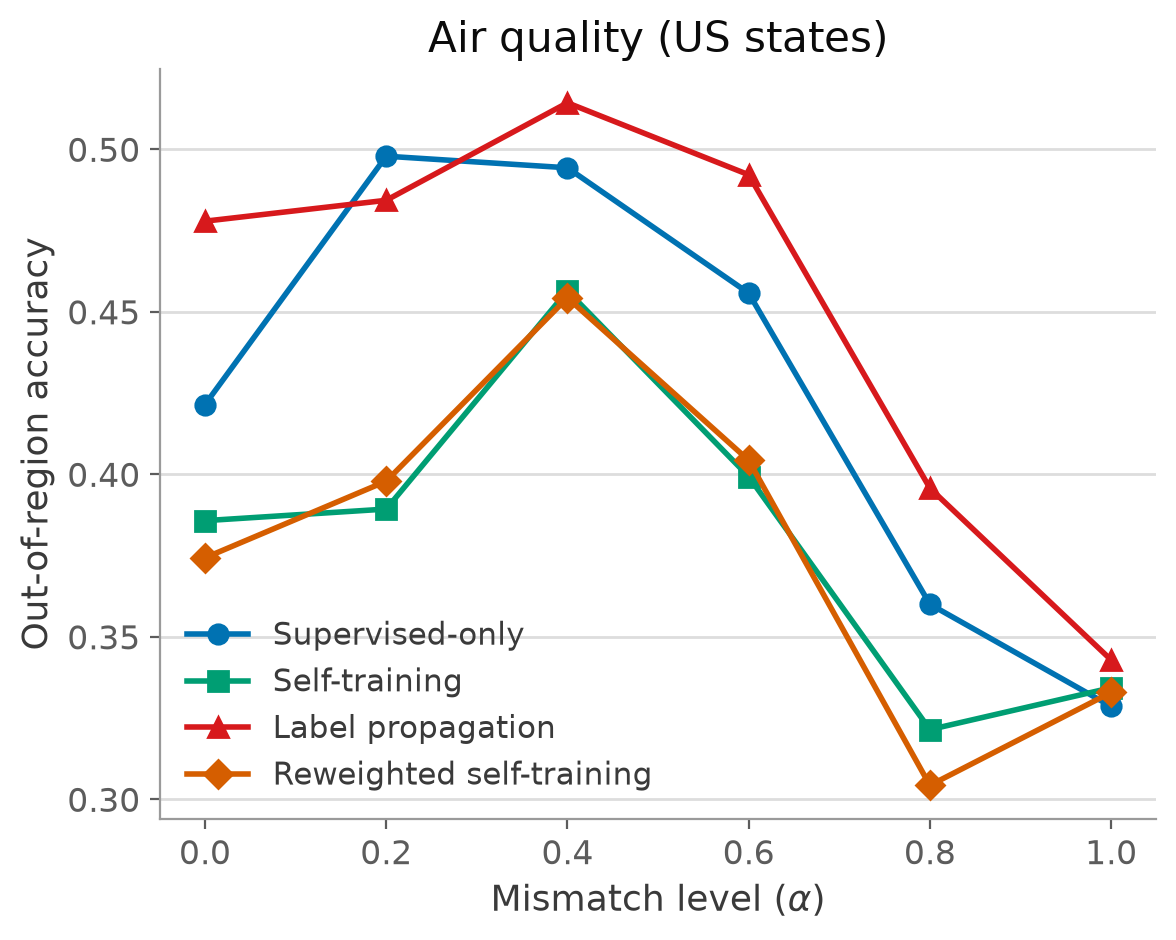}
    \caption{}
    \label{fig:real-c}
  \end{subfigure}
  \caption{Out-of-region accuracy as a function of mismatch severity for each real-world dataset, by
  method: (a) housing, (b) socio-economic, (c) air quality. Housing and air quality both show a
  late-onset decline consistent with the synthetic and WILDS pattern, whereas socio-economic
  accuracy increases with mismatch severity for self-training and re-weighted self-training only
  (Section~\ref{sec:res-real}).}
  \label{fig:real}
\end{figure}

The three real-world datasets were evaluated using the same mismatch-severity protocol as
PovertyMap-WILDS, with six severity levels, twenty random seeds, and four methods: the three
baseline approaches and the re-weighted self-training variant. Table~\ref{tab:t5} summarises
out-of-region accuracy at minimal and maximal mismatch severity, while
Figure~\ref{fig:real} shows the full performance curves. The results provide mixed support for H1 across the three datasets. The housing dataset most
closely reproduces the synthetic and PovertyMap-WILDS findings. Accuracy remains relatively
stable across most mismatch levels before declining sharply under the most severe restriction,
where labelled data are drawn only from Los Angeles County. Three of the four methods lose
between 12.2 and 13.8 percentage points, with confidence intervals excluding zero. Label
propagation is the exception, showing a smaller but still significant decline of 4.3
percentage points. The air quality dataset shows the same pattern with the largest declines of
the three real datasets: all four methods lose between 11.0 and 15.4 percentage points, every
95\% confidence interval excluding zero.

Socio-economic data does not follow this pattern, but the effect is narrower than a first
reading of Table~\ref{tab:t5} suggests. Out-of-region accuracy increases with mismatch severity
for self-training and re-weighted self-training (5.0 and 5.2 percentage points, both 95\%
confidence intervals excluding zero), but supervised-only and label propagation show no
reliable change (0.5 and 0.8 percentage points, both intervals including zero). This occurs
despite measured maximum mean discrepancy increasing over the same range, indicating that
greater distributional mismatch does not necessarily imply lower predictive performance for
methods that iteratively pseudo-label. To investigate this behaviour, we repeated the
experiment using a plain logistic regression trained on either the full labelled set or the
anchor-region-only labelled set, without self-training, pseudo-labelling, or re-weighting.
Unlike self-training and re-weighted self-training, this plain classifier showed almost no
change from restricting to the anchor region ($+0.3$ percentage points; Table~\ref{tab:t6}),
consistent with supervised-only and label propagation's null result above. The reversal is therefore not a generic property of the dataset, but appears specific to the two methods that iteratively pseudo-label the (always uniformly drawn) unlabelled pool. One plausible contributor is that the feature--label relationship transfers
unusually well across space here: the correlation vector computed within the anchor region
alone is almost identical to that of the held-out regions (cosine similarity 0.974), compared
with much weaker agreement for housing (0.786) and air quality (0.242) (Table~\ref{tab:t7}).
With transfer this strong, an anchor-only seed model should produce pseudo-labels about as
informative as an all-region seed model's, which would explain the \emph{absence} of a
mismatch penalty for these two methods; it does not by itself explain why accuracy
specifically increases, and we do not have a confirmed mechanism for that part of the effect.
A sample-size-matched random-region control (Supplementary Section~S1.3) shows this increase
is specific to the anchor region instead of a generic property of restricting to any single
region: self-training reaches 0.888 accuracy with the anchor at $\alpha=1$ but only 0.664 with
a randomly chosen region of the same labelled size, a 22.4-percentage-point difference. Strong
transfer explains why the anchor is not \emph{penalised}; it does not explain why this
particular anchor is unusually \emph{rewarded}, and we flag this as an open question rather
than a settled mechanism. These findings highlight that marginal distribution mismatch and
task difficulty are distinct
concepts, and that H1 should be interpreted as a general tendency instead of a universal
property; we report the socio-economic reversal as a genuine but method-specific
counter-example, not a property of the dataset that would appear under any SSL algorithm.

All four real-dataset results above use one fixed train/out-of-distribution region split (and,
for PovertyMap-WILDS, one fixed alphabetical country ordering). A reduced-cost check
(Supplementary Section~S1.4) against two alternative splits, and an anchor-distance country
ordering for PovertyMap-WILDS, shows the reported direction is not always stable: Housing's
decline shrinks to under one percentage point under one alternative split; socio-economic's
self-training reversal and air quality's decline each flip sign under at least one
alternative; and PovertyMap-WILDS's self-training decline reverses to an increase under
anchor-distance ordering. We report the specific split and ordering used throughout this
section as the ones fixed in advance for the main analysis, not as representative of every
possible split.

\begin{table}[t]
  \centering
  \caption{Replicating the $\alpha=0$-to-$\alpha=1$ reversal with a plain logistic regression (no
  self-training, pseudo-labelling, or re-weighting), trained on an all-region versus an
  anchor-region-only labelled seed and evaluated on each dataset's fixed out-of-region test set.
  Anchor share is the anchor region's percentage of the $\alpha=0$ labelled-candidate pool.}
  \label{tab:t6}
  \begin{tabular}{llccccc}
    \toprule
    Dataset & Anchor region & Anchor share (\%) & All-region acc. & Anchor-only acc. & $\Delta$ (pp) \\
    \midrule
    Socio-economic & Virginia     & 6.7  & 0.862 & 0.865 & $+0.3$ \\
    Housing        & Los Angeles  & 34.6 & 0.769 & 0.709 & $-6.0$ \\
    Air quality    & California   & 19.4 & 0.511 & 0.357 & $-15.4$ \\
    \bottomrule
  \end{tabular}
\end{table}

\begin{table}[t]
  \centering
  \caption{Cosine similarity between the feature--label correlation vector computed within the
  anchor region (or the full $\alpha=0$ training pool) and the same vector computed on the
  held-out out-of-region states, per dataset. Values close to 1 indicate the anchor region's
  covariate--label relationship matches the held-out regions'; values near 0 indicate it does not.}
  \label{tab:t7}
  \begin{tabular}{lcc}
    \toprule
    Dataset & cos(anchor, OOD) & cos(all-region, OOD) \\
    \midrule
    Socio-economic & 0.974 & 0.995 \\
    Housing        & 0.786 & 0.980 \\
    Air quality    & 0.242 & 0.176 \\
    \bottomrule
  \end{tabular}
\end{table}

\begin{table}[t]
  \centering
  \caption{Real-world datasets: out-of-region accuracy at minimal versus maximal mismatch severity, by
  method, averaged over seed, twenty seeds per level. Drop is reported with its 95\% seed-cluster
  bootstrap confidence interval.}
  \label{tab:t5}
  \begin{tabular}{llcccc}
    \toprule
    Dataset & Method & $\alpha=0$ & $\alpha=1$ & Drop (pp) & 95\% CI \\
    \midrule
    Housing & Self-training             & 0.736 & 0.598 & 13.8 & $[12.3,\,15.2]$ \\
    Housing & Supervised-only           & 0.779 & 0.654 & 12.5 & $[11.1,\,14.2]$ \\
    Housing & Re-weighted self-training & 0.728 & 0.606 & 12.2 & $[10.6,\,13.6]$ \\
    Housing & Label propagation         & 0.701 & 0.658 & 4.3 & $[2.3,\,6.1]$ \\
    \addlinespace
    Socio-economic & Re-weighted self-training & 0.842 & 0.894 & $-5.2$ & $[-6.8,\,-3.4]$ \\
    Socio-economic & Self-training             & 0.839 & 0.888 & $-5.0$ & $[-6.9,\,-3.1]$ \\
    Socio-economic & Supervised-only           & 0.839 & 0.844 & $-0.5$ & $[-1.8,\,0.6]$ \\
    Socio-economic & Label propagation         & 0.792 & 0.800 & $-0.8$ & $[-2.3,\,0.6]$ \\
    \addlinespace
    Air quality & Label propagation         & 0.525 & 0.371 & 15.4 & $[11.6,\,19.7]$ \\
    Air quality & Supervised-only           & 0.516 & 0.379 & 13.7 & $[10.1,\,17.6]$ \\
    Air quality & Re-weighted self-training & 0.511 & 0.386 & 12.6 & $[7.1,\,19.0]$ \\
    Air quality & Self-training             & 0.489 & 0.379 & 11.0 & $[5.9,\,17.7]$ \\
    \bottomrule
  \end{tabular}
\end{table}

\subsection{The distribution-aware framework}
\label{sec:res-framework}

The distribution-aware framework was evaluated as a secondary investigation into whether the
performance degradation caused by spatial mismatch can be mitigated. Re-weighting alone
provided only modest improvements over standard self-training. On PovertyMap-WILDS, the
reduction in performance loss was 2.7 percentage points in the twenty-seed experiments, and
across all five datasets the average performance drop decreased from 6.6 percentage points for
standard self-training to 5.6 percentage points.

Adding the geographic-proximity weighting and adaptive pseudo-labelling components did not
improve performance further. Instead, the complete framework produced a larger average
performance drop of 11.5 percentage points, indicating that these additional components were
not beneficial under severe spatial mismatch. A domain-adversarial variant consistently
outperformed the original framework on four of the five datasets.

As this mitigation study extends the paper's main contribution instead of defining it, the
complete framework, ablation analysis, and additional results are presented in Supplementary
Material Sections~S1.5 and~S2.1.

\subsection{Spatial awareness across method families}
\label{sec:res-comparative}

We further evaluated ten SSL methods spanning four methodological families to examine whether
the observed behaviour extends beyond the three baseline approaches. The complete comparison,
including per-dataset results, is provided in Supplementary Material Section~S2.2. The expanded evaluation provides only partial support for H3. Among the methods incorporating
spatial information, the spatial graph neural network showed a lower average endpoint drop than
the supervised-only baseline (4.9 vs.\ 6.9 percentage points), with one clear exception: on the
smallest dataset, air quality, its drop was more than double the supervised-only baseline's,
consistent with a $k$-nearest-neighbour graph built from too few points to support reliable
message passing. In contrast, the geostatistical method of
\citet{fouedjio2022geostatistical} no longer showed a clear robustness advantage in the
twenty-seed experiments, although it remained the strongest-performing method on the smallest
dataset. The broader comparison also shows that self-training is not uniquely representative of
pseudo-label-based methods. Modern consistency-based approaches, including FixMatch and Mean
Teacher, exhibited smaller average performance declines than self-training under increasing
spatial mismatch. The geographically weighted network remained the least robust method across
the evaluated settings. As this comparative study extends the paper's main contribution instead of defining it, the full
experimental results and method-specific analyses are presented in Supplementary Material
Section~S2.2.
 
\subsection{Sensitivity analysis and location of the breakdown}
\label{sec:res-sensitivity}

\begin{figure}[t]
  \centering
  \begin{subfigure}{0.49\linewidth}
    \includegraphics[width=\linewidth]{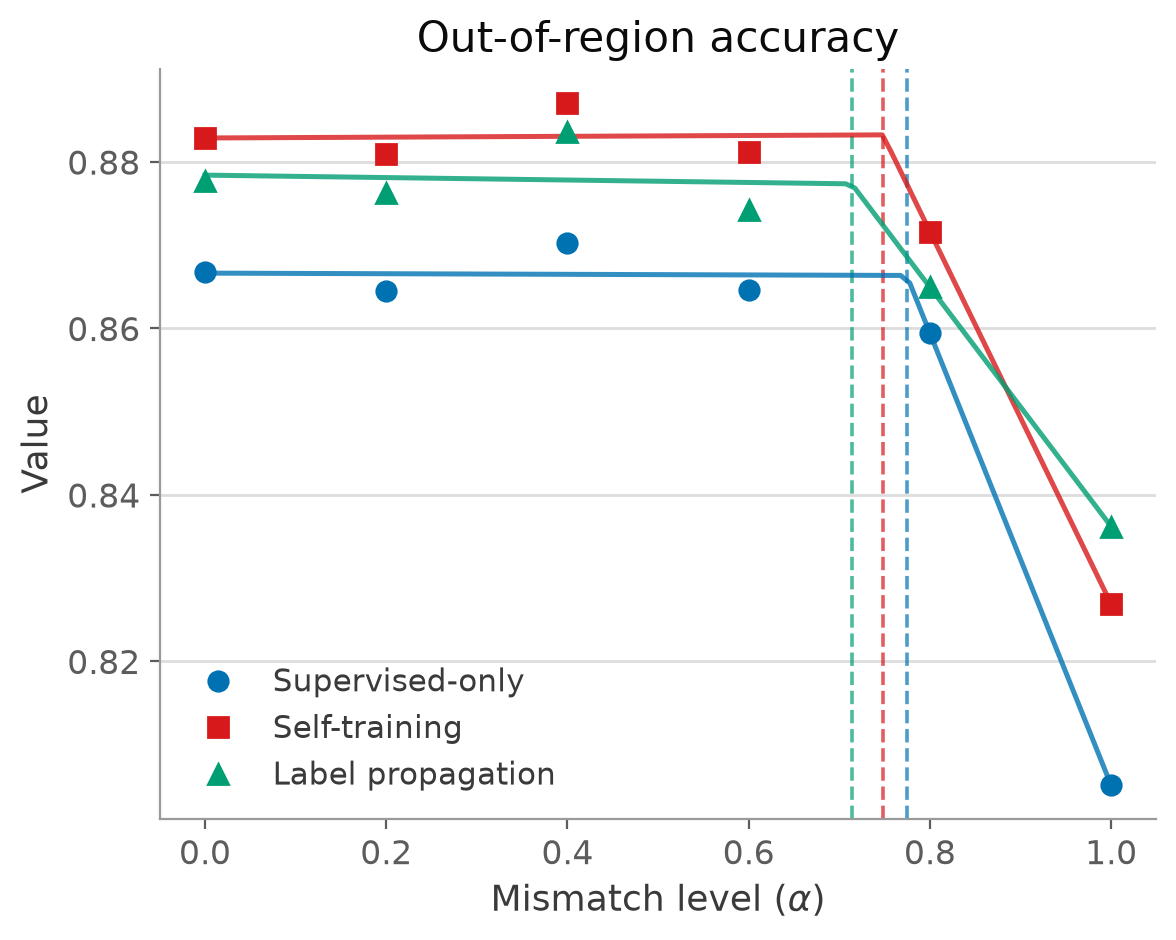}
    \caption{}
    \label{fig:changepoint-a}
  \end{subfigure}
  \hfill
  \begin{subfigure}{0.49\linewidth}
    \includegraphics[width=\linewidth]{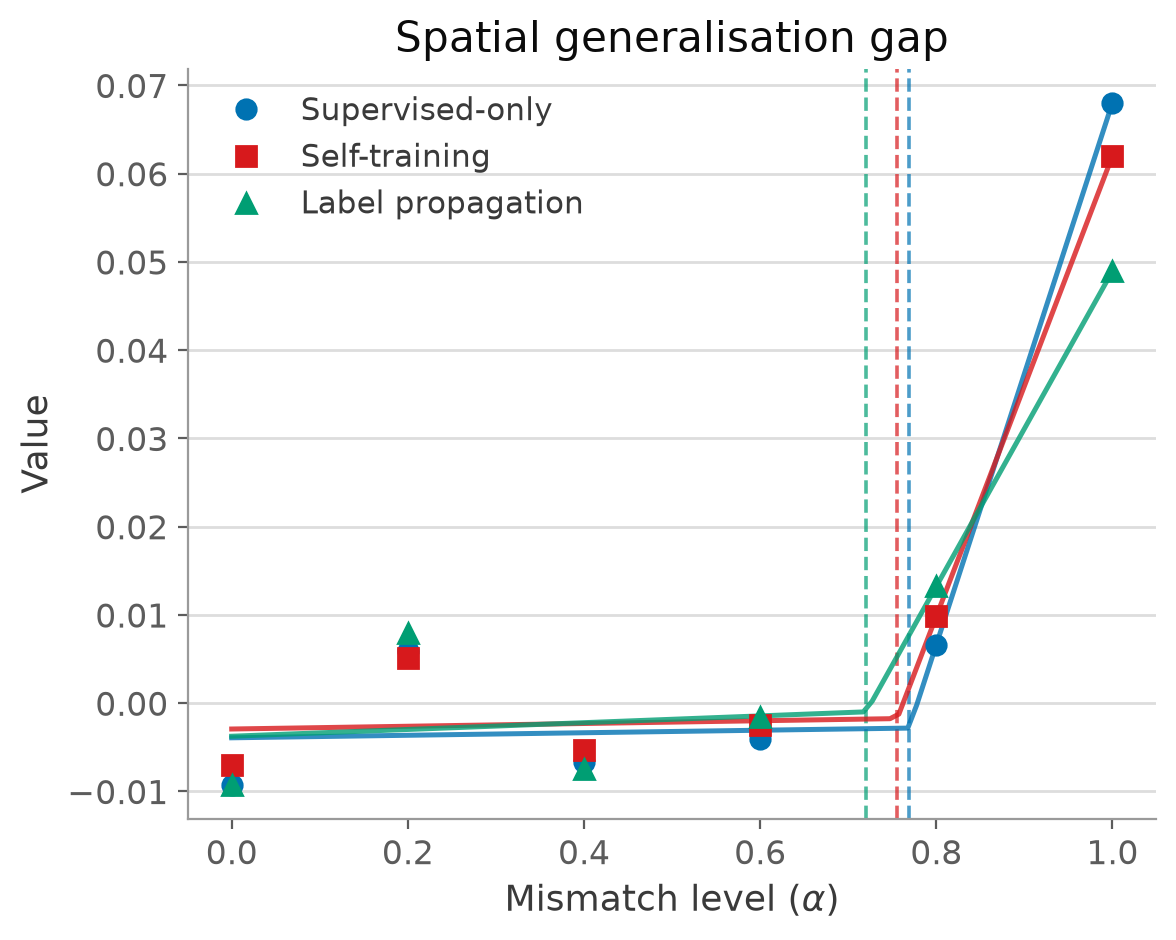}
    \caption{}
    \label{fig:changepoint-b}
  \end{subfigure}
  \caption{Fitted breakdown points (dashed vertical lines) for (a) out-of-region accuracy and
  (b) the spatial generalisation gap, by method, from the twenty-seed re-run.}
  \label{fig:breakpoints}
\end{figure}

The apparent breakdown around $\alpha \approx 0.6$ in
Figures~\ref{fig:acc-mismatch} and~\ref{fig:gap-mismatch} was formally analysed using the
two-segment piecewise-linear model in Equation~\eqref{eq:segmented}. The model was fitted
separately for each baseline method using out-of-region accuracy and the spatial
generalisation gap from the twenty-seed synthetic experiments (2,160 runs). Table~\ref{tab:t8}
summarises the estimated breakpoints with 95\% percentile bootstrap confidence intervals, and
Figure~\ref{fig:breakpoints} shows the fitted piecewise regressions.

Across methods and metrics, the estimated breakpoints range from 0.71 to 0.77, indicating that
the sharpest decline begins slightly later than suggested by visual inspection of the raw
performance curves. However, the precision of these estimates differs across methods. Five of
the six breakpoint estimates have relatively narrow confidence intervals, whereas the accuracy
breakpoint for label propagation is much less precise, with a 95\% confidence interval of
$[0.44,\,0.78]$. This broader interval is consistent with the more gradual decline already
observed for label propagation in Figure~\ref{fig:acc-mismatch}. The segmented regression analysis confirms that performance degradation is better
described as a threshold-like transition than as a gradual decline, while highlighting that the
location of this transition can be estimated more precisely for some methods than for others.

\begin{table}[H]
  \centering
  \caption{Fitted mismatch-severity breakpoints from the two-segment piecewise-linear model, with 95\%
  percentile bootstrap confidence intervals over 20 seeds. CI width is the interval's span. The last
  column is the BIC of the segmented fit minus the BIC of a single no-break linear fit to the same six
  points ($n=6$); more negative favours the segmented model, but see the caveat below the table.}
  \label{tab:t8}
  \begin{tabular}{llcccc}
    \toprule
    Method & Metric & Breakpoint & 95\% CI & CI width & $\Delta$BIC (seg. $-$ linear) \\
    \midrule
    Supervised-only   & Acc                    & 0.774 & $[0.733,\,0.925]$ & 0.19 & $-22.0$ \\
    Supervised-only   & $\Delta$   & 0.769 & $[0.735,\,0.798]$ & 0.06 & $-12.9$ \\
    Self-training     & Acc                     & 0.748 & $[0.689,\,0.788]$ & 0.10 & $-19.5$ \\
    Self-training     & $\Delta$   & 0.756 & $[0.700,\,0.863]$ & 0.16 & $-13.7$ \\
    Label propagation & Acc                     & 0.714 & $[0.438,\,0.776]$ & 0.34 & $-11.3$ \\
    Label propagation & $\Delta$   & 0.720 & $[0.599,\,0.777]$ & 0.18 & $-6.3$ \\
    \bottomrule
  \end{tabular}
\end{table}

In every method/metric pair, the two-segment model has substantially lower BIC than a single
no-break line fit to the same six mean-curve points, nominally favouring a break. We treat this
comparison as weak evidence instead of confirmation: with only six severity levels and five free
parameters (two slopes, an intercept, the breakpoint location, and the residual variance) against
one point per level, the segmented fit is close to interpolating the mean curve (residual sum of
squares within one to two orders of magnitude of zero at every row), which mechanically favours it
over any smooth alternative regardless of whether a true threshold exists. Distinguishing a genuine
break from a smooth, strongly nonlinear curve is not possible at this resolution and would require
denser sampling of $\alpha$, particularly between 0.6 and 1.0, which we leave to future work; the
$\alpha \approx 0.71$--$0.77$ estimates should accordingly be read as a threshold-like breakdown
specific to this synthetic generator, not a validated changepoint location.

\section{Discussion}

Marginal mismatch is a mechanism that degrades performance non-uniformly across methods and
non-gradually in severity, not a benign departure from SSL's assumptions.
The threshold-like breakdown, located between roughly $0.71$ and $0.77$ of this synthetic
generator's mismatch axis, is more informative than the existence of degradation
itself: spatial SSL may tolerate moderate sampling bias and then fail abruptly once bias
crosses a critical level. This location is specific to the synthetic generator and has not
been translated into an observed coverage or divergence threshold for real labelling
campaigns; doing so, and validating it outside the synthetic setting, is left to future work.
It complements covariate-space diagnostics such as the area of applicability
\citep{meyer2021aoa}. This extends the lesson of
\citet{oliver2018realistic}, that matched-distribution benchmarks flatter SSL, from an
in-distribution versus out-of-distribution binary to a continuously increasing mismatch, showing
the transition between safe and unsafe regimes to be sharp and precisely locatable. 

The comparison between methods paints a more nuanced picture than our original hypothesis
anticipated. In the synthetic experiments, self-training and supervised-only are not statistically
distinguishable once uncertainty is taken into account, while a paired seed-cluster bootstrap on
the between-method differences shows label propagation's smaller accuracy drop is reliably
distinguishable from both. The PovertyMap-WILDS point estimates and the calibration analysis
are directionally consistent with self-training being particularly vulnerable under severe
mismatch, though the WILDS accuracy ranking itself is not statistically distinguishable under
a paired bootstrap (Section~\ref{sec:res-wilds}); the calibration finding, that self-training
becomes increasingly overconfident outside the labelled region while label propagation
appears comparatively robust, is the more direct evidence of the two. The broader comparison across ten SSL methods adds a
further qualification: FixMatch, despite also relying on pseudo-labels, is substantially more
robust than conventional self-training. This suggests that pseudo-labelling alone is unlikely to
explain the observed behaviour, leaving the specific source of self-training's fragility as an
open question. 

The independent effect of spatial non-stationarity reinforces the distinction between
covariate shift and concept shift in the dataset-shift taxonomy of
\citet{morenotorres2012unifying}. Marginal mismatch changes the distribution of covariates,
whereas non-stationarity alters the relationship between covariates and the target. Their
comparable effects in isolation suggest that addressing covariate shift alone is unlikely to
eliminate performance degradation, a conclusion that is consistent with the modest gains
observed from the distribution-aware mitigation strategies. The divergence metrics evaluated in this study provide practical indicators of when spatial SSL is likely to become unreliable. Global divergence measures consistently tracked performance degradation, while the proposed kernel-weighted local estimator improved substantially over the
naïve fixed-grid approach by producing more stable local estimates under sparse sampling.
Together with the estimated breakdown threshold, these diagnostics offer a practical means of
assessing whether labelled data provide sufficient geographic coverage before deploying an SSL
model. 

The separation drawn in Section~\ref{sec:setup} between marginal mismatch and non-stationarity is
confirmed empirically here, not merely assumed: that non-stationarity degrades performance
when the labelled and unlabelled marginals are identical shows the two to be distinct mechanisms,
not one distribution-mismatch construct, exactly as the covariate-shift versus concept-shift
distinction of \citet{morenotorres2012unifying} predicts --- mismatch changes $p(x)$ with
$p(y \mid x)$ fixed, whereas non-stationarity changes $p(y \mid x)$ directly. Disentangling these
two, together with the autocorrelation channel that stresses SSL's independence assumptions,
not its distributional ones \citep{roberts2017crossval}, is itself part of the contribution, since
the semi-supervised literature routinely runs the three together. The separation also delimits what
the present methods can achieve: the validated divergence diagnostics and the located breakpoint
speak to the sampling-design failure --- marginal mismatch --- and the comparable effects of the two
mechanisms in isolation imply that a covariate-shift correction alone, as re-weighting is, can only
partially mitigate spatial SSL's failure modes. A complete treatment would need a concept-shift-aware
component that adapts locally in the manner of geographically weighted regression
\citep{fotheringham2002gwr} instead of only re-weighting which points are trusted, and the framework
results bear this out: re-weighting helps materially on PovertyMap-WILDS, close to a pure
covariate-shift manipulation, but not where non-stationarity or task-difficulty effects dominate.

\section{Conclusion}

Spatial data do not break semi-supervised learning; biased sampling does. Once that distinction
is drawn, the failure takes a clear and consistent shape. Marginal mismatch degrades performance
abruptly, through a threshold-like collapse located here between severities of $0.71$ and
$0.77$. The failure is easy to miss, since affected models grow confidently wrong exactly where
they are least reliable, so accuracy alone never raises the alarm. Which methods suffer varies, and not straightforwardly with the assumption each operationalises. Label propagation is reliably the most stable of the three baselines under synthetic mismatch, and self-training the most overconfident outside the labelled region; but the accuracy ordering does not reach significance on PovertyMap-WILDS, and FixMatch's robustness despite also relying on pseudo-labels leaves the source of self-training's fragility unresolved. And spatial non-stationarity degrades performance through a
mechanism entirely separate from mismatch, active even when the marginals are identical -- the
clearest evidence that marginal mismatch and non-stationarity are distinct mechanisms, and that
spatial dependence is not, in itself, the culprit.

These are not artefacts of a synthetic generator. 
The threshold breakdown and self-training's greater point-estimate fragility replicate on PovertyMap-WILDS with a larger effect size and recur across three real datasets -- two of them consistent with threshold degradation. Global divergence metrics track the loss more reliably than a naïve localised statistic, and a kernel-weighted correction raises the localised correlation from $-0.28$ to $-0.40$ --- closing part of the gap to the global measures. Mitigation is
the harder half of the story: density-ratio re-weighting helps only marginally even where covariate shift dominates, yet the fuller framework makes matters worse, and a model
leaning on geography alone is the least robust method in the study. The validated divergence
diagnostics are nonetheless usable today, as a relative indicator of reliability across
candidate labelled samples for the same deployment target, even though the specific breakpoint
located here has not been calibrated to an absolute, deployment-ready threshold -- while the
harder problem this leaves open, correcting concept shift instead of only covariate shift, is
where reliable spatial SSL will be won or lost. 

\clearpage

\end{document}